\let\TeXyear\year
\documentclass{IEEEaccess}
\let\setyear\year
\let\year\TeXyear

\usepackage{cite}
\usepackage{amsmath,amssymb,amsfonts}

\DeclareMathOperator*{\argmin}{arg\,min}
\usepackage{algorithm}
\usepackage[algo2e,ruled,vlined,linesnumbered]{algorithm2e}
\usepackage{graphicx}
\usepackage{textcomp}
\usepackage{hyperref}

\usepackage{cleveref}
\usepackage[font=small,labelfont=bf]{caption}
\usepackage{float}
\usepackage{multirow}
\usepackage{booktabs}

\usepackage{tikz}
\setyear{2022}

\newcommand*\circled[1]{\tikz[baseline=(char.base)]{
            \node[shape=circle,draw,inner sep=0.8pt, minimum size=2pt] (char) {#1};}}
\newcommand{\rpoint}[1]{\circled{{\fontfamily{pcr}\selectfont\footnotesize{#1}}}}

\SetKwInput{KwInput}{Input} 
\SetKwInput{KwOutput}{Output} 

\def\BibTeX{{\rm B\kern-.05em{\sc i\kern-.025em b}\kern-.08em
    T\kern-.1667em\lower.7ex\hbox{E}\kern-.125emX}}

\NewSpotColorSpace{PANTONE}
  \AddSpotColor{PANTONE} {PANTONE3015C} {PANTONE\SpotSpace 3015\SpotSpace C} {1 0.3 0 0.2}
  \SetPageColorSpace{PANTONE}%

\usepackage{xcolor}

\begin{document}
\history{Date of publication xxxx 00, 0000, date of current version xxxx 00, 0000.}
\doi{00.0000/ACCESS.0000.DOI}

\title{RoHNAS: A Neural Architecture Search Framework with Conjoint Optimization for Adversarial Robustness and Hardware Efficiency of Convolutional and Capsule Networks}

\author{\uppercase{Alberto Marchisio}\authorrefmark{1}, \IEEEmembership{Graduate Student Member, IEEE},
\uppercase{Vojtech Mrazek}\authorrefmark{2}, \IEEEmembership{Member, IEEE}, \uppercase{Andrea Massa}\authorrefmark{3}, \uppercase{Beatrice Bussolino}\authorrefmark{3}, \IEEEmembership{Graduate Student Member, IEEE}, \uppercase{Maurizio Martina}\authorrefmark{3}, \IEEEmembership{Senior Member, IEEE}, and \uppercase{Muhammad Shafique}\authorrefmark{4}, \IEEEmembership{Senior Member, IEEE}\vspace*{10pt}}
\address[1]{Embedded Computing Systems Group, Institute of Computer Engineering, Technische Universität Wien (TU Wien), Vienna, Austria}
\address[2]{Evolvable Hardware Research Group, Faculty of Information Technology, Brno University of Technology, Brno, Czechia}
\address[3]{VLSI Lab, Department of Electronics and Telecommunications, Politecnico di Torino, Turin, Italy}
\address[4]{eBrain Lab, Division of Engineering, New York University Abu Dhabi, UAE}
\tfootnote{This work has been supported in part by the Doctoral College Resilient Embedded Systems, which is run jointly by the TU Wien's Faculty of Informatics and the UAS Technikum Wien. This research is also partly funded through the NYUAD's Research Enhancement Fund (REF) Award on ``eDLAuto: An Automated Framework for Energy-Efficient Embedded Deep Learning in Autonomous Systems'', and partly supported by the NYUAD Center for Artificial Intelligence and Robotics (CAIR), funded by Tamkeen under the NYUAD Research Institute Award CG010. This work was also supported in part by Czech Science Foundation grant GA22-02067S. The computational resources were supported by the Ministry of Education, Youth and Sports of the Czech Republic through the e-INFRA CZ (ID:90140).}

\markboth
{Marchisio \headeretal: RoHNAS}
{Marchisio \headeretal: RoHNAS}

\corresp{Corresponding author: Alberto Marchisio (e-mail: alberto.marchisio@tuwien.ac.at).}

\begin{abstract}
Neural Architecture Search (NAS) algorithms aim at finding efficient Deep Neural Network (DNN) architectures for a given application under given system constraints. DNNs are computationally-complex as well as vulnerable to adversarial attacks. In order to address multiple design objectives, we propose \textit{RoHNAS}, a novel NAS framework that jointly optimizes for adversarial-robustness and hardware-efficiency of DNNs executed on specialized hardware accelerators. Besides the traditional convolutional DNNs, \textit{RoHNAS} additionally accounts for complex types of DNNs such as Capsule Networks. For reducing the exploration time, \textit{RoHNAS} analyzes and selects appropriate values of adversarial perturbation for each dataset to employ in the NAS flow. Extensive evaluations on multi - Graphics Processing Unit (GPU) - High Performance Computing (HPC) nodes provide a set of Pareto-optimal solutions, leveraging the tradeoff between the above-discussed design objectives. For example, a Pareto-optimal DNN for the CIFAR-10 dataset exhibits 86.07\% accuracy, while having an energy of 38.63 mJ, a memory footprint of 11.85 MiB, and a latency of 4.47 ms.
\end{abstract}


\titlepgskip=-15pt

\maketitle

\section{Introduction}\label{sec:introduction}

Among the Machine Learning algorithms, Deep Neural Networks (DNNs) have shown state-of-the-art performance in a wide variety of applications~\cite{Capra2020SurveyDNN}\cite{Grigorescu2019DLAD}. 
Finding an efficient DNN architecture for a given application through a Neural Architecture Search (NAS) is a very complex optimization problem, which involves a huge number of parameters and typically extremely long exploration time~\cite{Pham2018ENAS}. 
The search space becomes even bigger when employing NAS algorithms for new brain-inspired types of DNNs, such as the Capsule Networks (CapsNets)~\cite{Sabour2017DynRouting}. Such CapsNets, and more in general advanced DNN models, aim at providing high learning capabilities. 
However, these advancements in DNN architectures come with multiple design challenges:

\begin{enumerate}
\item \textit{High computational complexity:} DNNs need specialized hardware accelerators to be deployed and executed at the edge, where the resources are constrained~\cite{Marchisio2019DL4EC}.
\item \textit{Security:} DNN classifiers can be fooled by adversarial attacks, which are small and imperceptible perturbations added to the inputs~\cite{RobustML_shafique}. Such a threat is extremely dangerous for safety-critical applications~\cite{Cheng2018SafetyCriticalDNN}. Furthermore, integrating means for security during NAS is a challenging problem, but can enable robust DNN designs~\cite{Dave2022AgileDesignML}\cite{Shafique2021energyEfficientSecureEdgeAI}, as compared to the regular DNN design flow.
\end{enumerate}

Hence, the problem is: \textit{how to design complex DNNs in an energy-efficient and robust way through an automated multi-objective NAS framework?}

\subsection{Limitations of State-Of-The-Art and Scientific Challenges}

\begin{figure}[h]
    \centering
    \includegraphics[width=.98\linewidth]{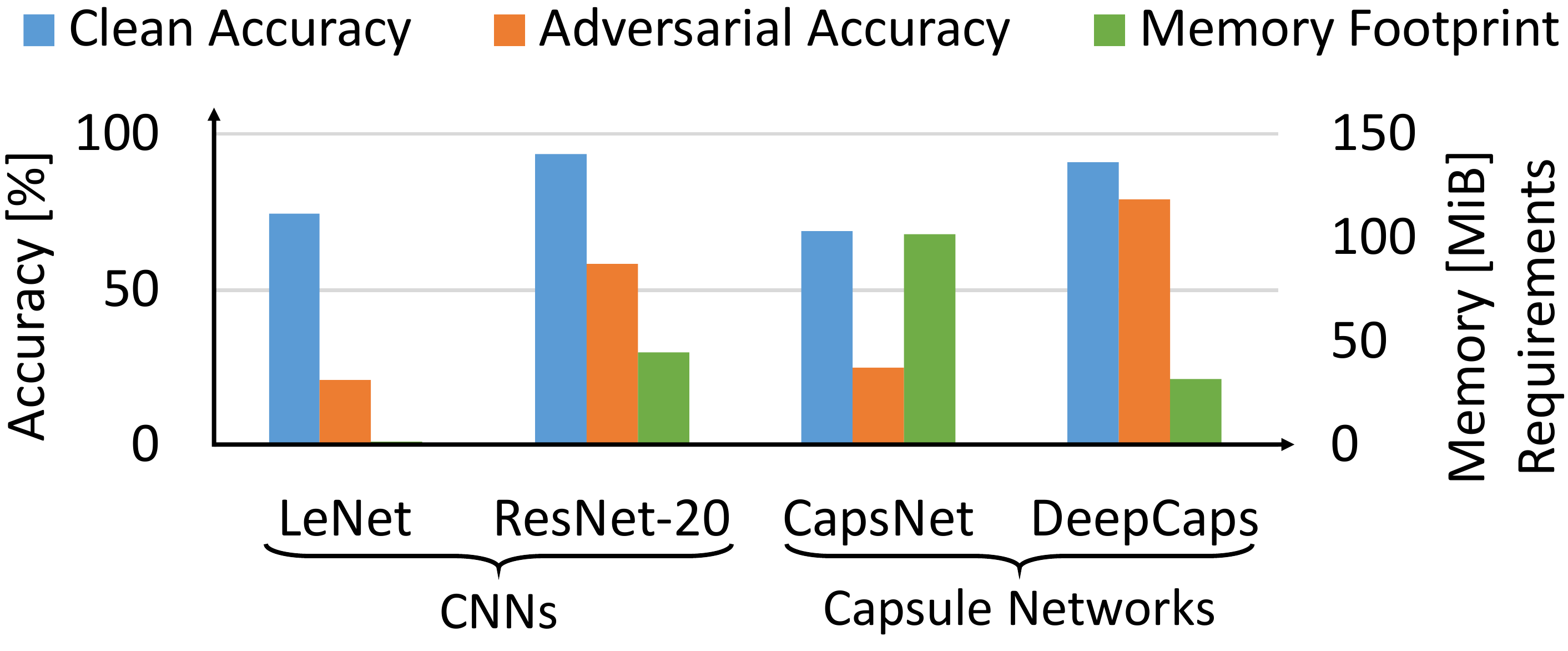}
    \caption{Adversarial robustness to the PGD attack vs. memory footprint of LeNet, CapsNet, ResNet-20, and DeepCaps for the CIFAR-10 dataset.}
    \label{fig:motivation_figure}
\end{figure}

Traditionally, the adversarial robustness of a given DNN is investigated a posteriori, i.e., once the DNN is already designed. The hardware efficiency of a DNN implemented on a given hardware accelerator is also a metric that is typically analyzed a posteriori, thus challenging the feasibility of its implementation on resource-constrained neuromorphic and/or IoT devices. 
We perform a motivational case study to analyze the adversarial accuracy\footnote{We refer to the \textit{adversarial accuracy} as the DNN test accuracy obtained when applying the adversarial attacks to every test example, i.e., by giving adversarial examples as input to the DNN.} and memory footprint of different DNNs, illustrating their adversarial robustness and complexity. We apply the Projected Gradient Descent (PGD) attack~\cite{Madry2017TowardsDL} with $\varepsilon=0.0001$ to the LeNet~\cite{Lecun1998MNIST}, the ResNet-20~\cite{He2016ResNet}, the CapsNet~\cite{Sabour2017DynRouting}, and the DeepCaps~\cite{Rajasegaran2019DeepCaps}, trained for the CIFAR-10\footnote{Performing numerous experiments for analyses and evaluation, constituting many NAS rounds on complex DNNs with CIFAR-10 dataset already took several weeks to months on our multi-GPU HPC node. Therefore, testing for bigger dataset is out of our currently available computational power and memory resource. Nevertheless, we believe that these findings are highly valuable, and would scale to bigger datasets as well.} dataset~\cite{CIFAR}. The results in Fig.~\ref{fig:motivation_figure} show that the LeNet~\cite{Lecun1998MNIST}, which is relatively small and shallow, is hardware efficient due to its low memory footprint, but relatively more vulnerable to attacks. A more complex DNN such as the ResNet-20~\cite{He2016ResNet} has a higher memory footprint but it also exhibits higher adversarial accuracy than the LeNet. Interestingly, the DeepCaps~\cite{Rajasegaran2019DeepCaps}, despite having a smaller memory footprint than the ResNet-20, is also relatively more robust against adversarial attacks. \textit{The goal of this paper is to integrate these diverse yet important objectives in a NAS framework to obtain Pareto-optimal solutions that explore the potential tradeoffs between different design objectives like computational complexity, memory, energy, latency, and/or security.}

Including the DNN security into the optimization goals of the NAS is a challenging task, because, besides the challenges in its representation in the design framework, it might lead to a massive search space explosion due to several additional factors and extremely time-consuming training and evaluations of numerous candidate solutions. A wide variety of adversarial attacks have been proposed in the literature~\cite{Yuan2019AdversarialExamplesSurvey}, and it is extremely complex to evaluate the adversarial robustness to different attack algorithms. A recent study in~\cite{Guo2020WhenNASMeetsRobustness} proposed a method evaluating the DNN robustness to the PGD attack~\cite{Madry2017TowardsDL} as the optimization goal of the NAS. \textit{On the contrary, our work performs joint optimizations for the adversarial robustness and hardware efficiency both, thereby leading to the increased complexity of the optimization problem, as well as large training time for evaluating the DNN robustness.} Moreover, it is challenging to model, implement and evaluate the hardware execution of different DNNs and CapsNets (including convolutional layers, fully-connected layers, and dynamic routing) in the NAS design flow.

\subsection{Our Novel Contributions}

\begin{figure}[h]
    \centering
    \includegraphics[width=\linewidth]{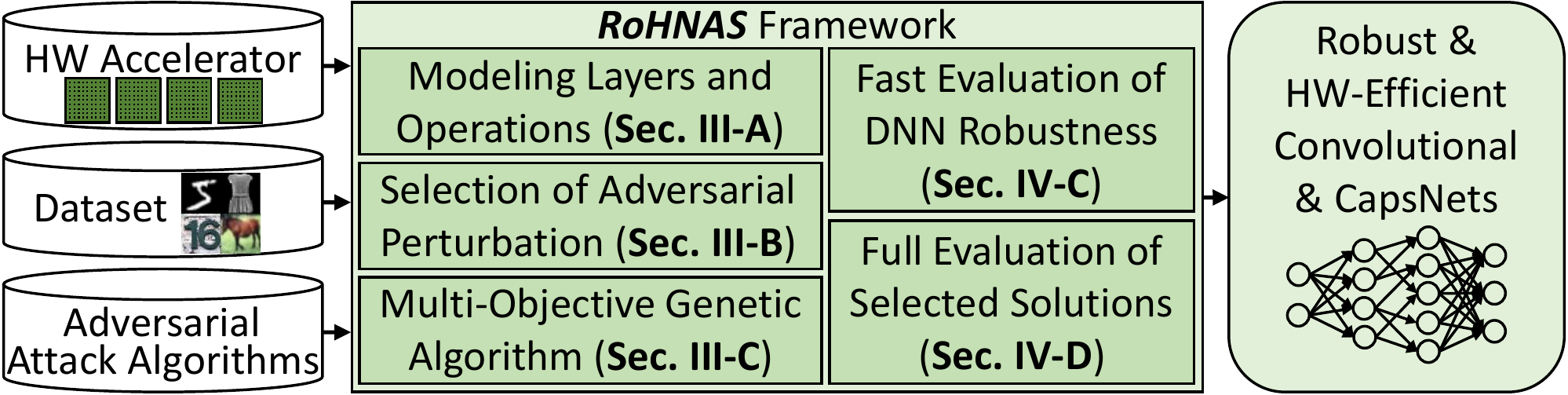}
    \caption{Overview of our \textit{RoHNAS} framework.}
    \label{fig:novel_contrib}
\end{figure}

To address the above-discussed challenges, we propose the novel \textit{RoHNAS} framework (see Figure~\ref{fig:novel_contrib}) that integrates multiple optimization objectives (like hardware efficiency and adversarial robustness) for diverse types of DNNs, like Convolutional Neural Networks (CNNs) and CapsNets. \textit{RoHNAS} employs the following key mechanisms:

\begin{enumerate}
    \item For architectural model flexibility and fast hardware estimation, we deploy analytical models of the layers and operations of DNNs and CapsNets, as well as their mapping and execution on specialized accelerators (\textbf{Section~\ref{subsec:analytical_model}}).
    \item To speed-up the robustness evaluation, we analyze and choose the values of the adversarial perturbations, which provide valuable differences when performing the NAS with DNNs subjected to such adversarial perturbations (\textbf{Section~\ref{subsec:significant_perturbation}}).
    \item We develop a specialized evolutionary algorithm, based on the principles of the Non dominated Sorting Genetic Algorithm II (NSGA-II) method~\cite{Deb2002NSGA-II}, to perform a multi-objective Pareto-frontier selection, with conjoint optimization for adversarial robustness, energy, memory, and latency of DNNs. (\textbf{Section~\ref{subsec:NSGA-II}})
    \item To reduce the overall training time, we devise a fast evaluation methodology for DNNs trained for a limited number of epochs (\textbf{Section~\ref{subsec:results_partially}}), while the Pareto-optimal solutions are evaluated after full-training, to obtain the exact results (\textbf{Section~\ref{subsec:results_fully}}).
\end{enumerate}

\noindent
\textbf{Implementation and Validation Contributions:} We have implemented our \textit{RoHNAS} using the TensorFlow library~\cite{tensorflow}, and evaluated more than 900 DNNs for the MNIST, Fashion-MNIST and CIFAR-10 datasets. Extensive validations are performed on Nvidia's multi-V100 Graphics Processing Unit (GPU) High Performance Computing (HPC) Nodes requiring weeks to months of experimentation time.

\noindent
\textbf{Open-Source Contribution:} For reproducible research, we release the code of the \textit{RoHNAS} framework at \url{https://github.com/ehw-fit/rohnas}.

\section{Background and Related Works}

\subsection{Adversarial Attacks}

DNNs are now deployed for a wide variety of applications, including safety-critical ones such as Autonomous Driving~\cite{Grigorescu2019DLAD}, Medicine~\cite{Barata2019DLCancer}, and Finance~\cite{Zanc2019FinanceDL}. Despite their performance, DNNs have severe security flaws, as adversarial attacks can fool DNNs with small input perturbations~\cite{RobustML_shafique}. Many studies~\cite{Szegedy2013IntriguingNN}\cite{Yuan2019AdversarialExamplesSurvey} have shown that DNNs are vulnerable to carefully crafted inputs designed to fool them. Very small imperceptible perturbations added to the data can completely change the output of the DNN model~\cite{RobustML2_shafique}.

It is essential for the attacker to \textit{minimize the added adversarial perturbation to avoid its detection}. 
Formally, given an original input $x$ with a target classification label $c$ with a DNN model $ m() $, the problem of generating an adversarial example $x^*$ can be formulated as a constrained optimization problem~\cite{Yuan2019AdversarialExamplesSurvey}:

\begin{equation}
\label{eq:adv}
     \begin{array}{rlclcl}
        x^* = \displaystyle \argmin_{x^*} & \mathcal{D}(x,x^*), \\
        s.t.~  & m(x) = c,  ~ m(x^*) = c^*,  ~ c \neq c^*
\end{array}
\end{equation}

Where $\mathcal{D}$ is the distance between two images and the optimization objective is to minimize this adversarial perturbation to make it stealthy. $x^*$ is considered as an adversarial example if and only if $ m(x) \neq  m(x^*) $ and the perturbation is bounded ($\mathcal{D}(x,x^*) < \epsilon $, where $\epsilon \geqslant 0 $).

Goodfellow et al.~\cite{Goodfellow2015FGSM} proposed the fast gradient sign method (FGSM) to generate adversarial examples by exploiting the gradient of the model w.r.t. the input images, towards the direction of the highest loss. Afterward, Madry et al.~\cite{Madry2017TowardsDL} and Kurakin et al.~\cite{Kurakin2016AdvExamplesPhysicalWorld}
proposed two different versions of the projected gradient descent (PGD) attack, an iterative version of the FGSM that introduces a perturbation $\alpha$ to multiple smaller steps. After each iteration, the PGD projects the generated image into a ball with a radius $\epsilon$, keeping the perturbation size small. 
It is a white-box attack and has both the targeted and untargeted versions. The algorithm consists of the following iteration:

\begin{equation}
     x^*_{i} = x^*_{i-1} - proj_{\varepsilon} (\alpha\cdot sign(\nabla_x loss(\theta,x,t)))
\end{equation}

Further details about different types of adversarial attacks and defenses can be found in comprehensive surveys such as~\cite{RobustML_shafique}\cite{Yuan2019AdversarialExamplesSurvey}. Moreover, recent works attempted to improve the DNN robustness against adversarial attacks by hash-based deep compression~\cite{Liu2018SecureDNNCompression} or approximate computing~\cite{Guesmi2021DefensiveApproximation}, thus requiring significant hardware design overhead.

\subsection{Convolutional and Capsule Network Hardware}\label{sec:capsaccmodel}

A wide variety of hardware architectures has been proposed for accelerating the execution of DNN inference~\cite{Sze2017EfficientDNN}\cite{Capra2020Updated}, focusing on improving the performance and energy-efficiency through compression, dedicated operation mapping, and specialized hardware design. Recently, hardware architectures for CapsNets have been proposed~\cite{Marchisio2021FEECA}. CapsNets layers require the execution of operations, such as dynamic routing, that are not supported by traditional DNN accelerators but crucial to detect changes in the compositional structure of the inputs~\cite{Venkatraman2020RoutingCompositional}.

\begin{figure}[h]
    \centering
    \includegraphics[width=.98\linewidth]{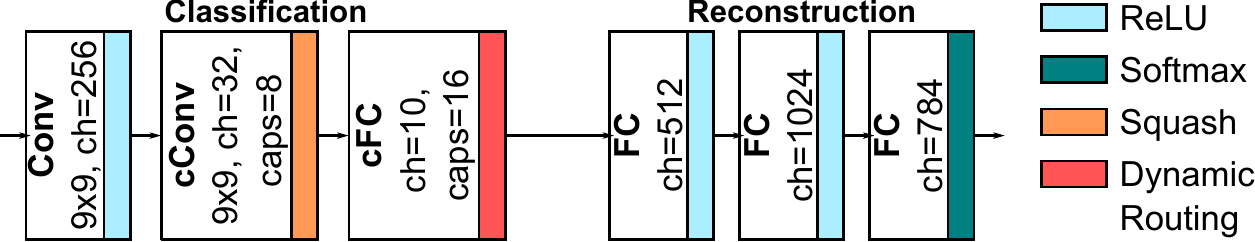}
    \caption{Architectural diagram of the CapsNet model of~\cite{Sabour2017DynRouting}.}
    \label{fig:capsnet}
\end{figure}

CapsNets, firstly proposed by the Google Brain's team~\cite{Hinton2011TransformingAE}, are elaborated DNN models in which the neurons are grouped together in vector form to compose the \textit{capsules}. Each neuron of a capsule encodes spatial information, while the vector's length encodes the probability of the entity being present. While the first architecture proposed in~\cite{Sabour2017DynRouting} is composed of only three layers, recently deeper CapsNet models were proposed~\cite{Rajasegaran2019DeepCaps}\cite{Sun2020DeepTensorCapsNet}. The main components of a CapsNet are the following:

\begin{itemize}
    \item \textbf{Convolutional (Conv) Layer:} The CapsNets need one or more traditional Conv layers to be applied at the beginning of the network. 
    \item \textbf{Convolutional or Fully-Connected (FC) Capsule Layers:} A generic CapsNet can contain some Conv capsule layers, whose principle of operation is identical to that of traditional Conv layers. However, the convolution is performed between the capsules rather than neurons, and the activation function needs to be the \textit{squash} operation (Eq.~\ref{eq:squash}), which constraints the length of the capsule vectors in the range [0,1].
    \begin{equation}\label{eq:squash}
        y = \frac{|x|^2}{(1+|x|)^2}\frac{x}{|x|}
    \end{equation}
    A CapsNet needs necessarily to be ended by a FC capsule layer, which mimics a traditional FC layer, but operating with capsules.
    \item \textbf{Dynamic Routing:} It is possible, but not necessary, to perform a \textit{dynamic routing} between two adjacent capsule layers. The dynamic routing~\cite{Sabour2017DynRouting} is an iterative algorithm which associates \textit{coupling coefficients} to the capsules predictions. The coupling coefficients of the capsules predicting the same result with greater confidence are maximized.
\end{itemize}
Figure~\ref{fig:capsnet} shows a simple three-layers CapsNet, as presented in~\cite{Sabour2017DynRouting}. CapsNets need to be completed by a \textit{reconstruction network}, consisting of FC layers or Transposed Conv layers, to reconstruct the input image.

\begin{figure}[h]
    \centering
    \includegraphics[width=\linewidth]{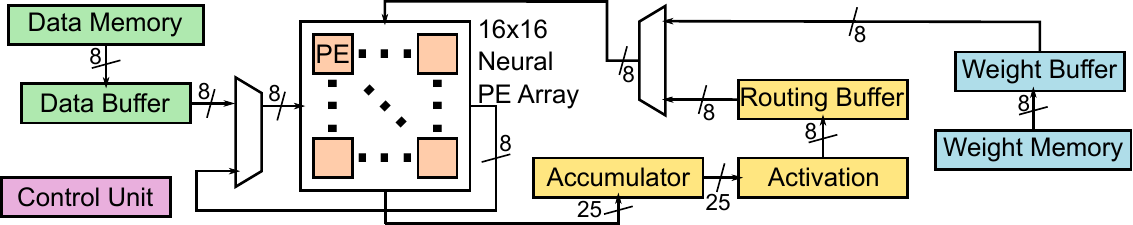}
    \caption{Architectural diagram of the CapsAcc accelerator of~\cite{Marchisio2019CapsAcc}.}
    \label{fig:capsacc}
\end{figure}

The aforementioned CapsAcc~\cite{Marchisio2019CapsAcc} accelerator (see Figure~\ref{fig:capsacc}) has been proposed to efficiently deploy the CapsNets in hardware, adapting to the specific needs of the dynamic routing and the squash operation. CapsAcc differs from other DNN accelerators for its Activations unit, which can apply the squash function in addition to the traditional Rectified Linear Unit (ReLU) and Softmax functions. Moreover, a Routing Buffer is inserted to store the partial results generated during the execution of the dynamic routing algorithm. Dedicated scratchpad memories are employed to minimize the energy consumption at runtime~\cite{Marchisio2021DESCNet}.

\subsection{Hardware-Aware NAS and Robust NAS}

Traditional NAS algorithms~\cite{Pham2018ENAS}~\cite{NAS_RL}~\cite{single_path_nas} have aimed at finding a high accurate DNN model for a given task, i.e., the DNN model which provides the highest accuracy on a given dataset. For example, the Efficient Neural Architecture Search (ENAS) algorithm~\cite{Pham2018ENAS} has generated a new architecture with 55.6 perplexity on the Penn Treebank~\cite{penntree} dataset. Recently, the interest in hardware efficiency has been growing, leading to designing Hardware-Aware NAS (HA-NAS) methodologies~\cite{Sekanina2021NASSurvey}. The main difference between traditional NAS and HA-NAS algorithms is that the latter also consider the hardware-deployment efficiency of candidate models, e.g., in terms of energy consumption, latency, or memory footprint. Among the related works, there exist mainly three types of heuristic search algorithms for the HA-NAS, which are (1) evolutionary algorithms, (2) reinforcement learning, and (3) differentiable NAS. The Accuracy-and-Performance-aware Neural Architecture Search (APNAS)~\cite{Achararit2020APNAS}, which is based on reinforcement learning, extends the ENAS algorithm by including the performance of DNNs executed in hardware in the optimization objectives of the NAS. AttentiveNAS~\cite{Wang2021AttentiveNAS} jointly optimizes the DNNs' accuracy and the computational complexity in terms of Mega FLoating Point Operations (MFLOPs). MnasNet~\cite{mnasnet} takes as an objective the inference latency and measures it by executing the candidate models on mobile phones. In~\cite{NAS_155}, an extended search space is used, which includes architecture parameters, quantization, and hardware parameters, precisely the tiling factors. Targeting the Field Programmable Gate Arrays (FPGAs), the FPGA-implementation aware Neural Architecture Search (FNAS) algorithm~\cite{fnas} uses an analytical model to consider the latency only. HotNAS~\cite{hotnas} targets energy efficiency by including model compression in the search space and supporting hardware for compressed models. During the candidate selection, the Single Path One-Shot (SPOS) NAS~\cite{spos} applies latency and FLoating Point Operations (FLOPs) constraints. HURRICANE~\cite{hurricane} generates a search space tailored to a specific hardware platform, considering the FLOPs and number of parameters, and their effect on the latency. The Differentiable NAS (DNAS) framework~\cite{Wu2019FBNet}, in which the search space is represented by a stochastic super net, explores a layer-wise space where each layer of the CNN corresponds to a different block, and the learning is conducted by training the super net. These works are primarily for the CNN models, and cannot handle Capsule Networks. 

On the other hand, recent works have also proposed NAS methodologies to achieve high robustness against adversarial attacks. In~\cite{Guo2020WhenNASMeetsRobustness}, a \textit{supernet} containing all the possible architectures in the search space is trained. Then, subnetworks are sampled from the supernet and evaluated in terms of accuracy and robustness to adversarial attacks. In~\cite{shashank_nas}, the search space is expanded to include some combinations of layers that have been proven to be particularly effective against adversarial attacks. However, all the works that focus on NAS for adversarial attacks have not yet considered the hardware efficiency aspects as conjoint optimization objectives. Moreover, these works are primarily for the CNN models, and cannot handle CapsNets. Recently, NASCaps~\cite{nascaps} has proposed a NAS methodology for CapsNets based on an evolutionary algorithm, but it cannot handle robustness challenges, and does not explore the tradeoffs between hardware efficiency and adversarial robustness.

\textit{Our RoHNAS framework distinguishes from the previous works because it combines for the first time hardware efficiency and robustness to adversarial perturbations as joint optimization goals for the NAS, and targets both CNN and CapsNets models.}

\section{RoHNAS Framework}
\label{sec:framework}

\begin{figure}[h]
    \centering
    \includegraphics[width=\linewidth]{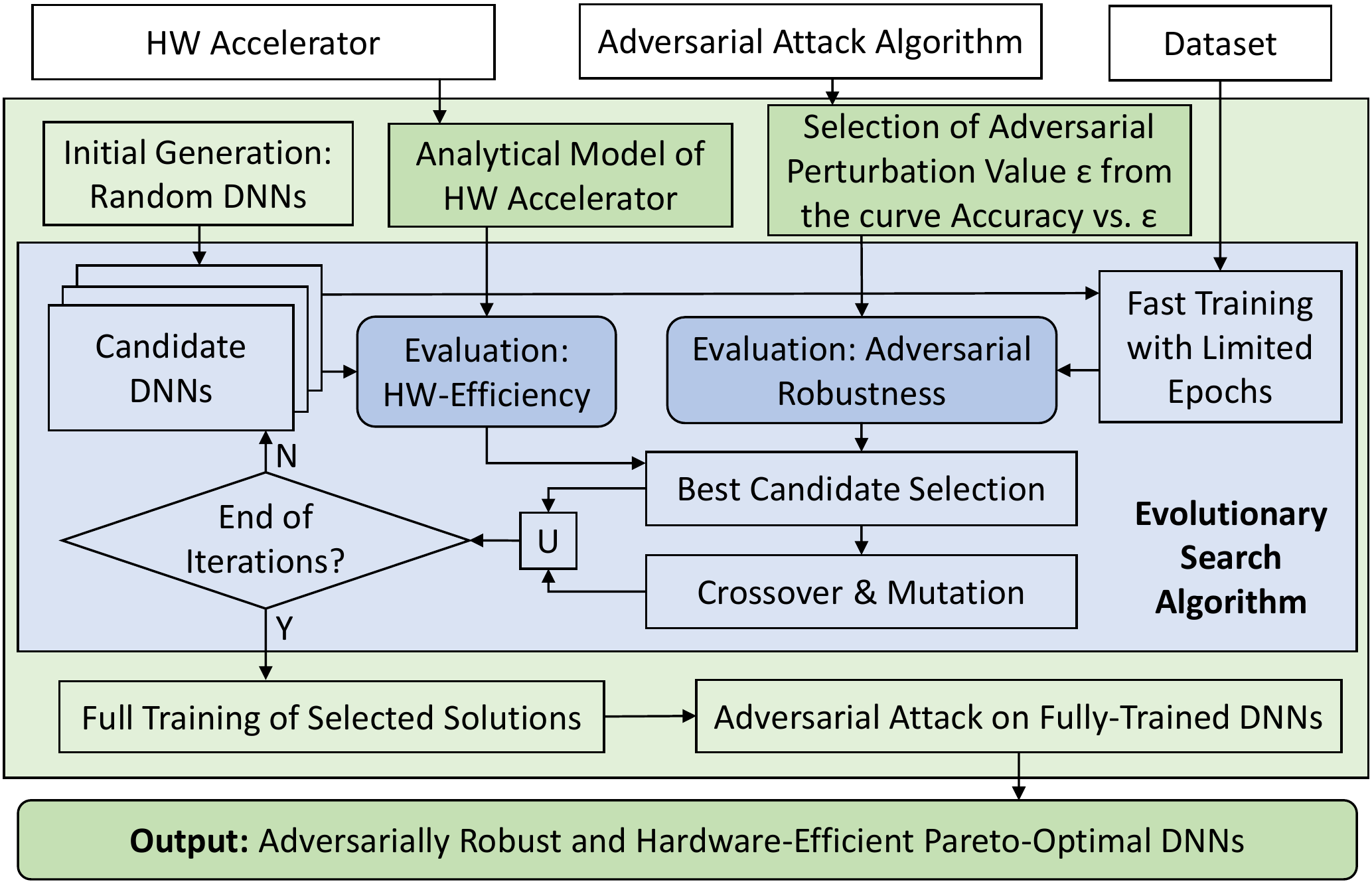} 
    \caption{Overview of our \textit{RoHNAS} framework and its key functionalities.} 
    \label{fig:framework_overview}
\end{figure}

Our evolutionary algorithm-based NAS methodology performs a multi-objective search. It automatically searches for inherently robust yet hardware-efficient DNN models by selecting Pareto-optimal candidates in terms of robustness, energy, latency, and memory footprint. The search space comprises both CNNs and CapsNets. The workflow of our \textit{RoHNAS} framework is shown in Figure~\ref{fig:framework_overview}, and is explained in detail in the following subsections. 

The framework's inputs are the hardware accelerator, the algorithm for generating the adversarial attack, and the dataset. After modeling analytically the hardware accelerator, the appropriate values of the adversarial perturbation to employ in the search are selected. This process, as will be described more in detail in \Cref{subsec:significant_perturbation}, consists of analyzing the accuracy vs. adversarial perturbation curve, and focusing on the high variation region which corresponds to the highest slope of the curve. After selecting the values of the adversarial perturbation to employ in the search, the evolutionary search algorithm (based on the principles of the NGSA-II genetic algorithm~\cite{Deb2002NSGA-II}) performs an iterative exploration through crossover, mutation, and best DNN candidate selection based on the objectives. To speed up the process, during the evolutionary algorithm, the adversarial robustness is evaluated after a fast training, i.e., for DNNs trained with a limited number of epochs, where its number is determined based on the Pearson Correlation Coefficient~\cite{PearsonCoefficient}. Towards generating exact robustness results, the set of Pareto-optimal DNN models are fully-trained, and the robustness against the adversarial attack on fully-trained DNNs is evaluated.

\subsection{Layer and Operation Modeling}
\label{subsec:analytical_model}

\begin{figure}[h!]
    \centering
    \includegraphics[width=.95\linewidth]{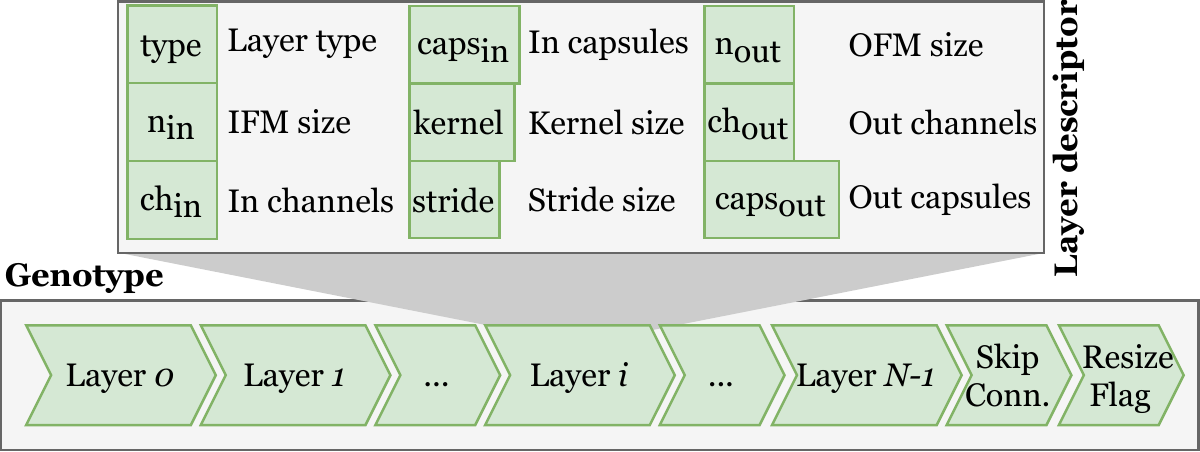} 
    \caption{Genotype structure. IFM stands for Input Feature Map, while OFM means Ouput Feature Map.} 
    \label{fig:gene}
\end{figure}

The \textit{RoHNAS} framework models each layer through a \textit{layer descriptor}, which contains all the relevant architectural parameters necessary to describe a generic DNN layer using a position-based representation. As shown in Figure \ref{fig:gene}, a layer descriptor contains all the information to construct its related layer, such as layer type, input feature map (IFM) size, input channels, input capsules, kernel size, stride size, output feature map (OFM) size, output channels, and output capsules. Using these parameters, it is possible to build many different types of CNN or CapsNet layers. Moreover, such a modular representation can easily be extended to support different layer types.
Multiple layer descriptors, together with information on extra skip connections and resizing of the inputs, form a \textit{genotype}, which allows describing various CNN and CapsNet architectural models.

To estimate the execution requirements of a DNN model on a specialized DNN hardware accelerator (e.g., CapsAcc~\cite{Marchisio2019CapsAcc} or Tensor Processing Unit (TPU)~\cite{Google2017TPU}), it is necessary to know its underlying hardware characteristics, for instance:

\begin{itemize}
    \item $T$, the clock period; 
    \item $Load\_Weights$, the number of clock cycles necessary to load the weights into the Processing Element (PE) array; 
    \item $P_{PEarray}$, the power consumed by the PE array, here estimated with the Synopsys Design Compiler tool;
    \item $E_{memory}$, the energy required for one memory access, here estimated with the CACTI-P tool~\cite{Li2011CACTI-P};
\end{itemize}


Knowing these parameters makes it possible to estimate the latency, energy consumption, and memory footprint of a DNN model analytically. From the dimensions of the layers of a given DNN, we assess: 
\begin{itemize}
    \item $w_l$, the number of weights in a layer; 
    \item $s_l$, the number of values to be summed to obtain an output value, for each layer; 
    \item $f_l$, the number of feature maps to be multiplied by the same weight, for each layer;
    \item $c_l$, the number of clock cycles needed to process a layer.
\end{itemize}
Given the model and the hardware features, the number of groups of weights loaded into the array ($w_{PEarray}$) and the number of memory accesses ($m\_acc$) can be determined through Equations~\ref{eq:wloads} and~\ref{eq:memory}, respectively. By computing the clock cycles (see Eq.~\ref{eq:cycles}), it is possible to estimate the latency and energy consumption, which, in conjunction with the memory footprint, form the set of hardware parameters computed through Eq.~\ref{eq:hwparameters}.

\begin{equation}
\label{eq:wloads}
w_{PEarray} = \left\lceil \frac{ w_l } { 16  \cdot  \min{(16, s_l)}}  \right\rceil 
\end{equation}

\begin{equation}
\label{eq:memory}
m\_acc = \begin{cases}
 256, \ \ \ \ \ \ \ \ \ \ \ \ \ \ \ \ \ \  \ \text{if } f_l = 1  \\ 
 16 \cdot \max(s_l - 15, 1),\ \text{otherwise}
 \end{cases}
\end{equation}

\begin{equation}
\label{eq:cycles}
c_l = w_l \cdot w_{PEarray}+f_l
\end{equation}

\begin{equation}\label{eq:hwparameters}
\resizebox{.9\linewidth}{!}{
$\begin{gathered}
latency = \sum_{l \in L}{c_l \cdot T}\\
energy = \left\lceil \frac{m\_acc}{128}  \right\rceil \cdot E_{memory} + \sum_{l \in L}{c_l \cdot T \cdot P_{PEarray}}\\
memory\ footprint = \sum_{l \in L}{w_l}
\end{gathered}$
}
\end{equation}

The model has been validated by comparing the results with the hardware implementation of the CapsAcc~\cite{Marchisio2019CapsAcc}. Recent studies in~\cite{nascaps} have also shown that the above-discussed parameters and such analytical models are sufficient to accurately estimate the latency, energy, and memory footprint of a given DNN model. In the following, we discuss the efficacy of our analytical models by comparing the estimated values with the real values of latency, energy, and memory requirements. By comparing our analytical model with the real implementation of the CapsNet~\cite{Sabour2017DynRouting} on CapsAcc~\cite{Marchisio2019CapsAcc}, our model provides accurate estimations of latency and memory footprint, and underesimates the energy consumption by around 25\%. Such a difference might be due to other elements of the
hardware implementation (e.g., interconnection overhead) that are not considered by the analytical model. Despite this underestimation, the fidelity of our models is high, i.e., all candidates have similar underestimation trend, so the selection of the candidates would not be affected by this underestimation of analytical models. Please note that our main focus was to have fast estimation with high fidelity.

\subsection{Design Space Reduction by Selecting an Appropriate Adversarial Perturbation Value}
\label{subsec:significant_perturbation}

Since the design space can potentially explode by considering several types and strengths of adversarial perturbations, the \textit{RoHNAS} framework restricts the design space by automatically selecting the values of adversarial perturbations to be used in the NAS for a given dataset. Algorithm~\ref{algo:epsilon} summarizes the proposed procedure. For each element of the testing dataset, the adversarial example is generated through the PGD algorithm~\cite{Madry2017TowardsDL} (line 4). Note, here we use PGD for illustrative reasons, and other adversarial attack algorithms can be integrated into our \textit{RoHNAS} framework. The parameter $\varepsilon$ determines the amount of adversarial perturbation. When considering the variation of the accuracy w.r.t. $\varepsilon$, as we will show in Section~\ref{subsec:impact_epsilon}, the region in which the slope is highest is in the middle of the graph, which corresponds to half of the clean accuracy, i.e., $\frac{Acc_0}{2}$ when considering that $Acc_0$ is the clean accuracy. By exploiting this intuition, our algorithm selects $\varepsilon_{NAS}$, which is the value of adversarial perturbation that provides the closest accuracy to the desired value of $\frac{Acc_0}{2}$. The selected value of $\varepsilon_{NAS}$ is employed in the \textit{One EPS} search, which optimizes for the robustness against one value of perturbation. Moreover, aiming at covering a wider spectrum of adversarial perturbation range, the \textit{Two EPS} search is devised. $\varepsilon_{low}$ and $\varepsilon_{high}$ are selected (lines~10-11), and the NAS is conducted by optimizing for the adversarial accuracy with both values.

\begin{algorithm2e}[h]
\SetAlgoLined
\KwInput{Deep Neural Network: $N$; \\
Test Dataset: $\mathcal{D} = \bigcup\limits_j X_j$; \\
Adversarial Perturbation Budget: $\varepsilon_i \in \mathcal{E} = [\varepsilon_{MIN}, \varepsilon_{MAX}]$;}
\KwOutput{Perturbation to apply for the NAS: $\varepsilon_{NAS}$;}
\BlankLine

$Acc_0 = Accuracy(N(D))$\;
\For{$i \in < \mathcal{E} >$}{
     \For{$j \in < \mathcal{D} >$}{
        $X'_{ij} = PGD (N, \varepsilon_i, X_j)$\;
    }
    $\mathcal{D}'_i = \bigcup\limits_j X'_{ij}$\;
    $Acc_i = Accuracy(N(\mathcal{D}'_i))$\;
}
$\varepsilon_{NAS} = \varepsilon_i\ :\ Acc_i \approx \frac{Acc_0}{2}$\;
$\varepsilon_{low} \approx \frac{\varepsilon_{NAS}}{10}$\;
$\varepsilon_{high} \approx 3 \cdot \varepsilon_{NAS}$\;

\caption{Adversarial Perturbation Selection.} \label{algo:epsilon}
\end{algorithm2e}

\subsection{Multi-Objective Evolutionary Algorithm}
\label{subsec:NSGA-II}

The selection of the Pareto-optimal solutions in the \textit{RoHNAS} framework is based on the principles of the NSGA-II algorithm~\cite{Deb2002NSGA-II}. The main core of the search algorithm is iterated for $g$ times, where each iteration $g_i$ represents a \textit{generation}. Each generation $g_i$ consists of a set of \textit{parent candidates} $P_i$, from which the \textit{offspring candidates} $Q_i$ are generated. 

At each generation $g_i$, the offsprings are generated from the parents via \textit{crossover} and \textit{mutation}. To perform the crossover operation, two parents $P_a$ and $P_b$ are randomly selected from the whole set of parent candidates. The genotypes of $P_a$ and $P_b$ are then pseudo-randomly splitted in two parts, obtaining four genotypes: $P_{a,1}$, $P_{a,2}$, $P_{b,1}$ and $P_{b,2}$. Two offsprings are then obtained concatenating the four genotypes as follows: 

\begin{equation}
    Q_a = P_{a,1} \& P_{b,2} \hspace{1cm} Q_b = P_{b,1} \& P_{a,2}
\end{equation}

To perform a mutation, a random parameter of a random layer descriptor is selected and modified. In particular, the kernel size, the stride, the skip connections, and the number of output capsules can be affected. When the generation of the offsprings is complete, it is necessary to check the validity of the solutions and in case remove the invalid candidates. 

\begin{figure}[h]
 \centering
 \includegraphics[width=.75\linewidth]{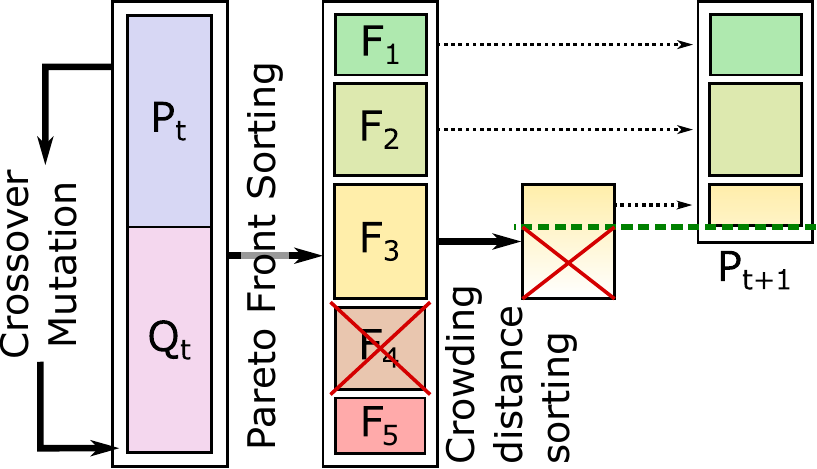}
 \caption{One iteration of the NSGA-II algorithm.}
 \label{fig:nsga}
\end{figure}

After the crossover and mutation processes, the set of candidates is the union of the parents and the offsprings sets. To select the best candidates, that will then be the parents in the next generation, the solutions are divided into a series $F_1, F_2, ..., F_N$ of Pareto-fronts, where $F_1$ is the best Pareto-front. The next-generation parents' set $P_{i+1}$ is filled with the solutions from the best Pareto-front. To obtain the chosen number of candidates, it may be necessary to select only a certain number of solutions from a Pareto-front (e.g., $F_3$ in Figure~\ref{fig:nsga}). In this case, the Pareto-front's solutions are sorted by \textit{crowding distance}, and the best ones are picked. 

\section{Evaluation of the RoHNAS Framework}
\label{sec:results}

\subsection{Experimental Setup}

\begin{figure}[h]
    \centering
    \includegraphics[width=\linewidth]{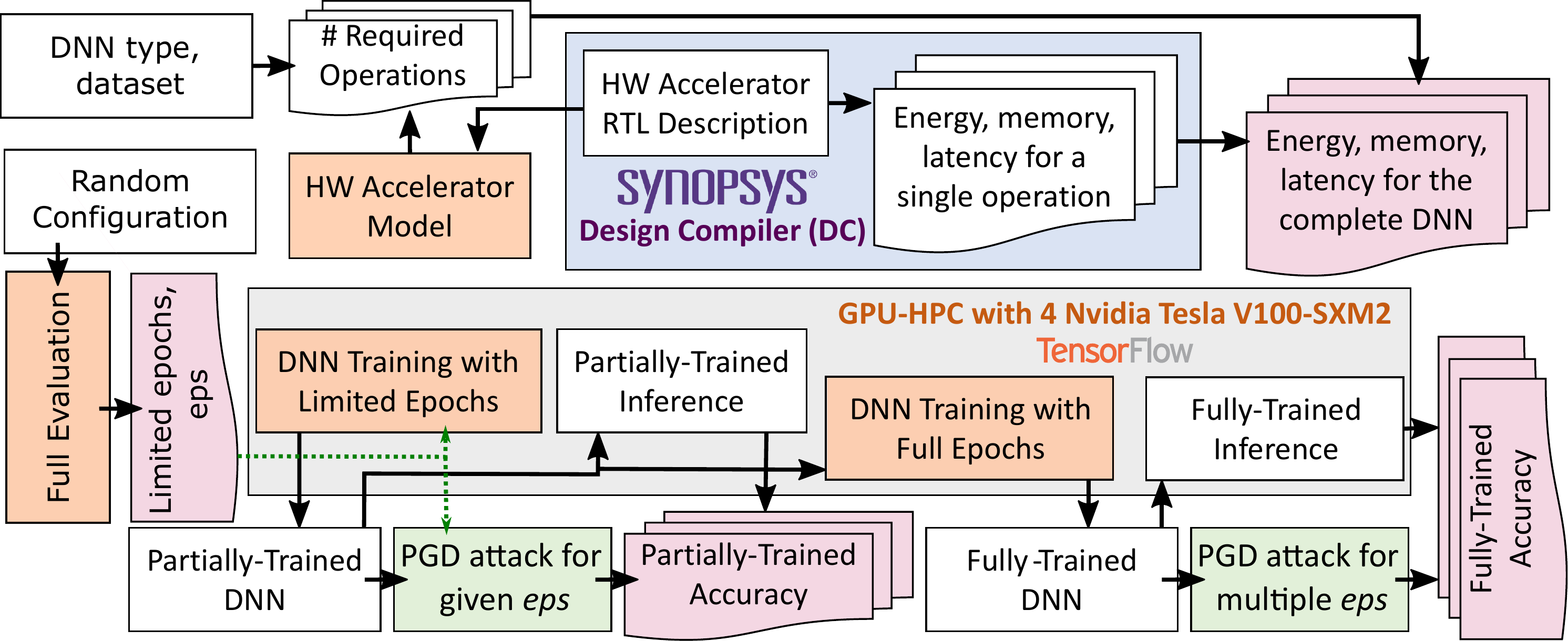}
    \caption{Tool-flow and setup for conducting the experiments.}
    \label{fig:exp_setup}
\end{figure}

The flow of our experiments and the tools used to implement the \textit{RoHNAS} framework are summarized in Fig.~\ref{fig:exp_setup}. The PGD adversarial attack algorithm~\cite{Madry2017TowardsDL} has been implemented with the CleverHans library~\cite{CleverHans}. The hardware model has been implemented using the open-source NASCaps library~\cite{nascaps}, which is based on the CapsAcc architecture~\cite{Marchisio2019CapsAcc} synthesized using the Synopsys Design Compiler tool, with a 45nm technology node and a clock period of 3ns. The training and testing of the DNNs, implemented in TensorFlow~\cite{tensorflow} have been running on the GPU-HPC computing nodes equipped with four NVIDIA Tesla V100-SXM2 GPUs. Note that, our experiments were running for 2,000 GPU hours with our fast evaluation method and 8,000 GPU hours for the final training and PGD attack evaluation. Without such exploration time reductions, or by considering more complex optimization problems (e.g., larger datasets or deeper DNN models), the exploration time would have lasted several GPU months.

The search algorithm is initialized with a random population of 10 elements, running for a maximum of 20 iterations of the genetic loop. The offspring population size is 10, and the mutation probability is 10\%. Each convolutional layer can be composed of a $3 \times 3$, $5 \times 5$, or $9 \times 9$ kernel, with a stride of either 1 or 2. The channels and capsule dimensions can both span between 1 and 64.

\subsection{Selection of Adversarial Perturbation for the NAS}
\label{subsec:impact_epsilon}

\begin{figure}[h]
    \centering
    \includegraphics[width=\linewidth]{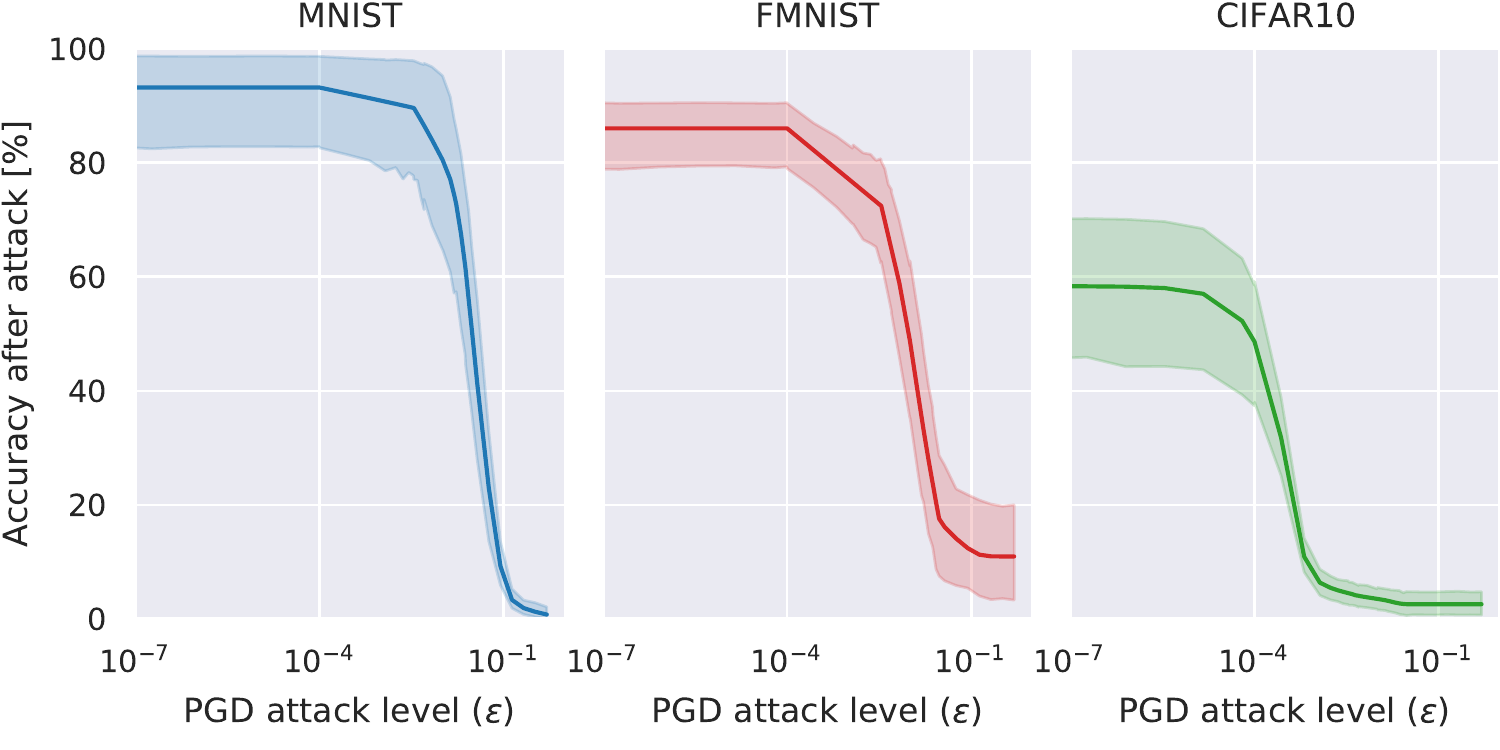}
    \caption{Analysis of the DNN robustness under the PGD attack, with different adversarial perturbation values, for MNIST, Fashion-MNIST, and CIFAR-10.}
    \label{fig:eps_analysis}
\end{figure}

The amount of adversarial perturbation is a key parameter to be selected for performing the NAS. Following the procedure described in Section~\ref{subsec:significant_perturbation}, the Pareto-optimal DNNs of the NASCaps library~\cite{nascaps} have been tested under the PGD attack~\cite{Madry2017TowardsDL}, with different values of the adversarial perturbation~$\varepsilon$. The results reported in Fig.~\ref{fig:eps_analysis} show that, as expected, the higher $\varepsilon$ is, the lower the DNNs' accuracy drops. The selected values for the NAS are reported in Table~\ref{tab:epsilon}. The selection process follows the procedure described in \Cref{algo:epsilon}. The \textit{One EPS} column refers to the search using a single value of $\varepsilon$, while the \textit{Two EPS} column refer to a search conducted with two different values of $\varepsilon$, which are called $\varepsilon_{low}$ and $\varepsilon_{high}$. Note, a simple dataset like the MNIST requires a relatively high adversarial perturbation to impact the DNN robustness. On the other hand, on a more complex dataset like the CIFAR-10, a smaller perturbation is already sufficient to misclassify a certain set of inputs.

\begin{table}[h]
\centering
\caption{Selected values of the adversarial perturbation $\varepsilon$ for the NAS, for MNIST, Fashion-MNIST and CIFAR-10 datasets. There are also reported the values of $\varepsilon_{low}$ and $\varepsilon_{high}$ for the \textit{Two EPS} search, which will be used for comparison in Section~\ref{subsec:results_fully}.}
\begin{tabular}{c|c|c|c}
& Two EPS $\varepsilon_{low}$ & One EPS $\varepsilon$ & Two EPS $\varepsilon_{high}$ \\
\hline
MNIST    & 3e-3             & 3e-2        & 1e-1              \\
F-MNIST  & 1e-3             & 1e-2        & 3e-2              \\
CIFAR-10 & 3e-5             & 3e-4        & 1e-3             
\end{tabular}
\label{tab:epsilon}
\end{table}

\subsection{RoHNAS Results with Fast DNN Robustness Evaluation}
\label{subsec:results_partially}

As discussed in Section~\ref{sec:framework}, to reduce the exploration time, our algorithm trains the DNNs only for a limited number of epochs, which results in a fast robustness evaluation. The similarity w.r.t. the full-training robustness has been measured through the Pearson Correlation Coefficient~\cite{PearsonCoefficient}, using the procedure described in~\cite{nascaps}. The choice of 10 training epochs for the CIFAR-10 dataset and 5 epochs for the Fashion-MNIST and MNIST datasets leverages the tradeoff between a high correlation and low training time.

\begin{figure}[h!]
    \centering
    \includegraphics[width=\linewidth]{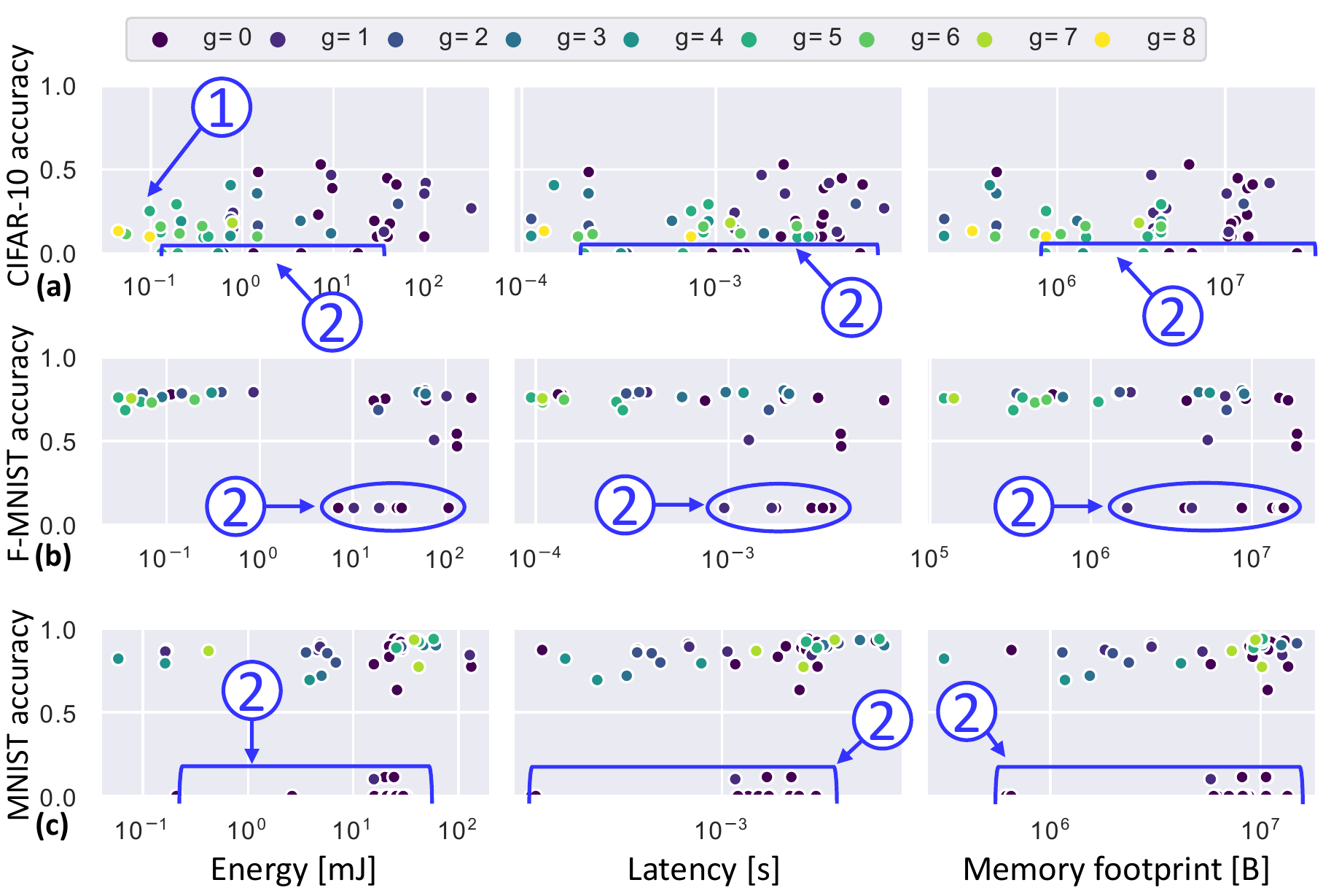}
    \caption{\textit{RoHNAS}' fast evaluation of DNN robustness under PGD attack, showing tradeoffs w.r.t. energy, latency and memory footprint. (a) Results for CIFAR-10. (b) Results for Fashion-MNIST. (c) Results for MNIST.}
    \label{fig:nas_attack_partial}
\end{figure}

The results of the \textit{RoHNAS - One EPS} with fast robustness evaluation are reported in Fig.~\ref{fig:nas_attack_partial}. The earliest generation of the algorithm produces sub-optimal DNN solutions, while most Pareto-optimal solutions are found in the latest generation. Note that, for the \textit{RoHNAS} evaluated on the CIFAR-10 dataset, the latest generations find DNNs that are less robust to the PGD attack, but still belong to the Pareto-frontier due to the low energy consumption (see pointer~\rpoint{1}). Note that, as highlighted by pointer~\rpoint{2}, several candidate DNNs found in the earliest generations are highly vulnerable to the PGD attack and are automatically discarded by the Pareto-frontier selection.

\subsection{RoHNAS Exact Results for Pareto-Optimal DNNs}
\label{subsec:results_fully}

\begin{figure}[h!]
    \centering
    \includegraphics[width=\linewidth]{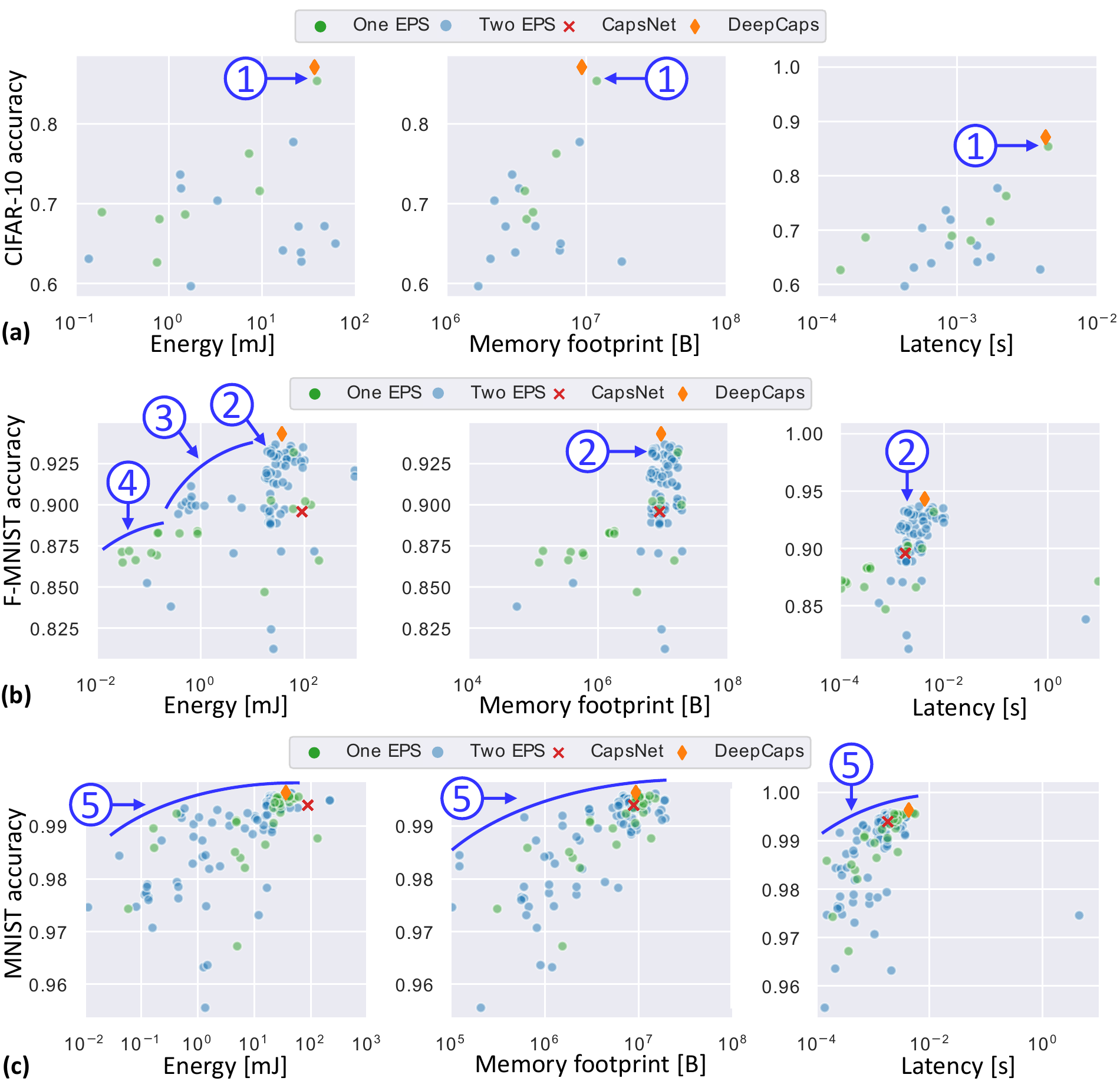}
    \caption{\textit{RoHNAS}' exact robustness evaluation of Pareto-optimal DNN solutions under the PGD attack, showing tradeoffs w.r.t. hardware-efficiency. (a) Results for CIFAR-10. (b) Results for Fashion-MNIST. (c) Results for MNIST.}
    \label{fig:nas_attack_full}
\end{figure}

The Pareto-optimal DNNs that are selected at the previous stage have been \textit{fully-trained} to obtain an exact robustness evaluation. The DNNs for the MNIST and Fashion-MNIST datasets have been trained for 100 epochs, while 300 epochs of training has been used for the DNNs targeting the CIFAR-10 dataset. The results reported in Fig.~\ref{fig:nas_attack_full} show tradeoffs between the design objectives. As highlighted by pointer~\rpoint{1} in Fig.~\ref{fig:nas_attack_full}, a Pareto-optimal solution found by the \textit{RoHNAS} framework for the CIFAR-10 dataset achieves 86.07\% accuracy while having an energy consumption of 38.63 mJ, a memory footprint of 11.85 MiB, and a latency of 4.47 ms. Similarly, the solution for the Fashion-MNIST dataset pointed in~\rpoint{2} reaches an accuracy of 93.40\% while having 6.40 ms latency, 61.19 mJ energy, and 16.82 MiB memory. Note that, while the \textit{Two EPS} search finds Pareto-optimal solutions in the middle range of energy (see pointer~\rpoint{3}), other interesting low-energy solutions are found by the \textit{One EPS} search, as indicated in pointer~\rpoint{4}. The Pareto-optimal DNNs's search for MNIST covers a more heterogeneous range of values, leveraging tradeoffs between different objectives (see pointer~\rpoint{5}).

\begin{figure}[h!]
    \centering
    \includegraphics[width=.98\linewidth]{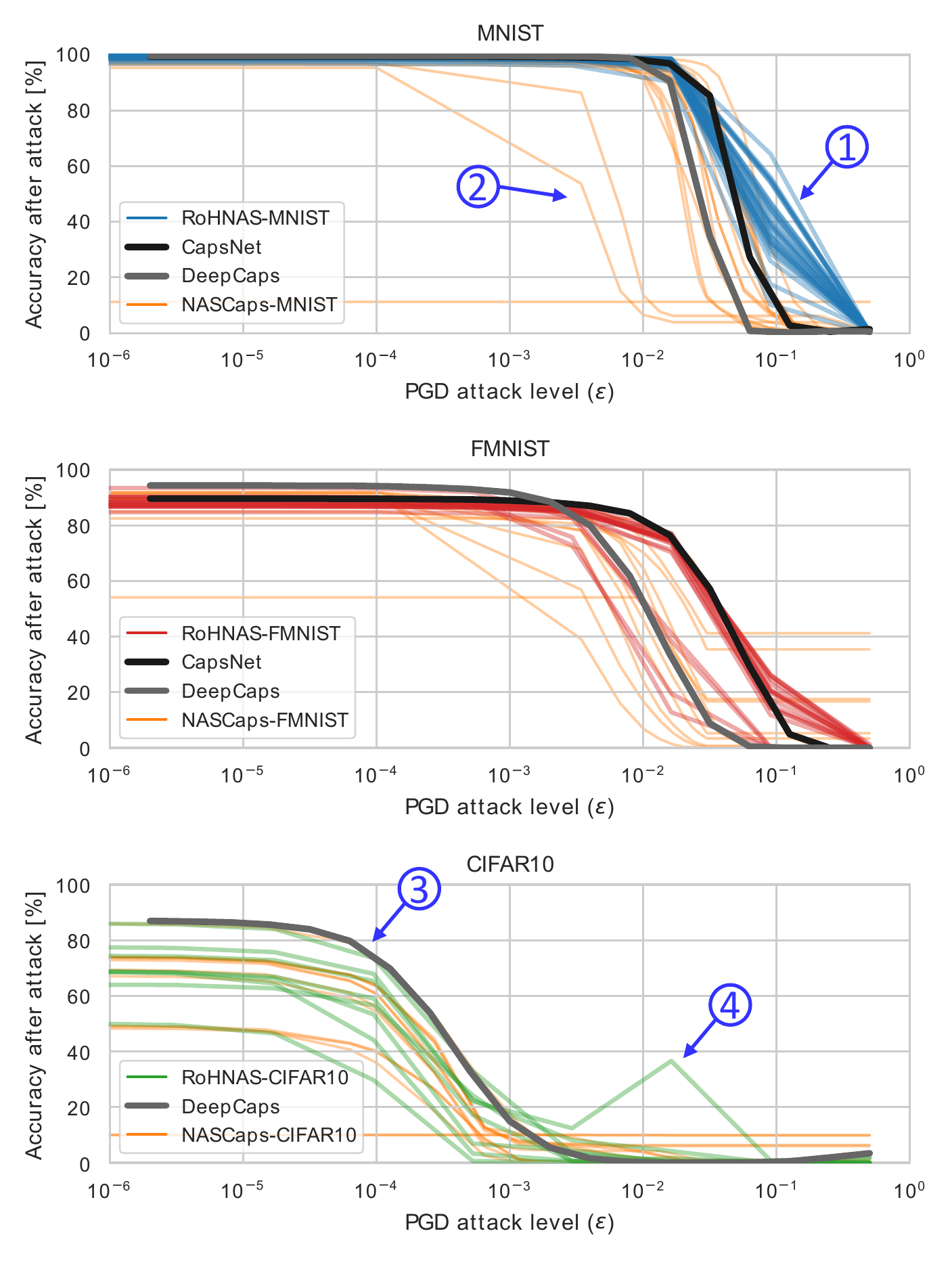}
    \caption{Evaluation of the \textit{RoHNAS} framework with the \textit{One EPS} setting, compared to other state-of-the-art architectures and NAS algorithms.}
    \label{fig:attack_oneeps_pointer}
\end{figure}

The \textit{RoHNAS} framework has been compared with other state-of-the-art DNN and CapsNet architectures, and NAS methodologies that include capsule layers in the search space. Fig.~\ref{fig:attack_oneeps_pointer} shows the comparison between our \textit{RoHNAS} framework (\textit{One EPS} setting), NASCaps~\cite{nascaps}, CapsNet~\cite{Sabour2017DynRouting} and DeepCaps~\cite{Rajasegaran2019DeepCaps}. For the MNIST dataset, the Pareto-optimal solutions generated with the \textit{RoHNAS} framework are particularly robust for a high range of perturbation $\varepsilon$ (see pointer~\rpoint{1}). Indeed, the accuracy starts dropping at around one order of magnitude higher $\varepsilon$ than NASCaps (see pointer~\rpoint{2}). For the Fashion-MNIST, the robustness behavior of the Pareto-optimal DNNs selected with the \textit{RoHNAS} frameweork is closely related to the CapsNet. Instead, for the CIFAR-10 dataset, the \textit{RoHNAS} DNNs' behavior is similar to the DeepCaps for low values of $\varepsilon$ (see pointer~\rpoint{3}), while a Pareto-optimal \textit{RoHNAS} solution offer a respectable robustness also with higher adversarial perturbation (see pointer~\rpoint{4}).

\begin{figure}[h!]
    \centering
    \includegraphics[width=.98\linewidth]{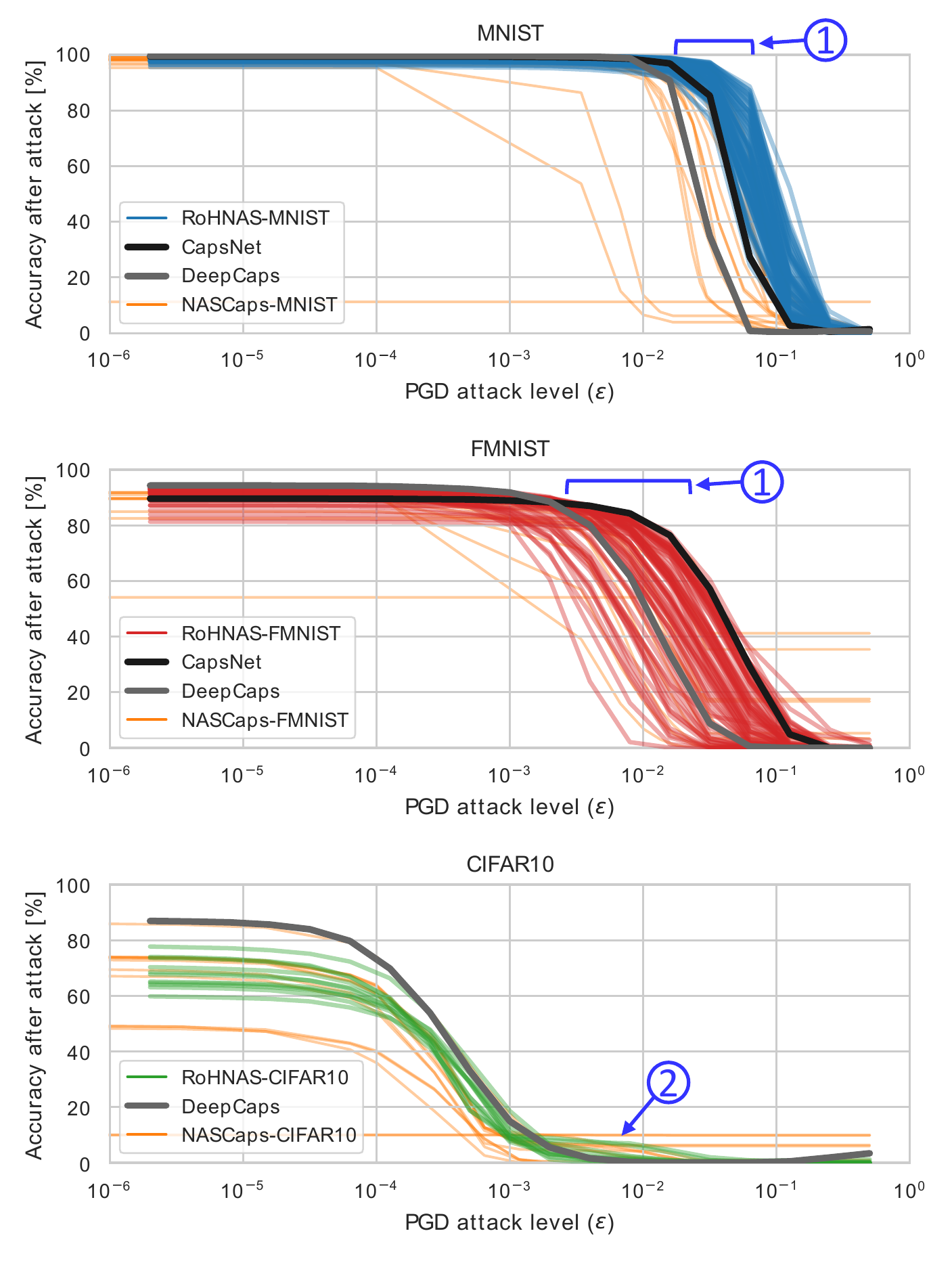}
    \caption{Evaluation of the \textit{RoHNAS} framework with the \textit{Two EPS} setting, compared to other state-of-the-art architectures and NAS algorithms.}
    \label{fig:attack_twoeps_pointer}
\end{figure}

The evaluation of the \textit{RoHNAS} framework with the \textit{Two EPS} setting is shown in Fig.~\ref{fig:attack_twoeps_pointer}. Compared to the \textit{One EPS} setting, the NAS produces different levels of robustness w.t.r. $\varepsilon$ for the MNIST and Fashion-MNIST datasets (see pointer~\rpoint{1} in Fig.~\ref{fig:attack_twoeps_pointer}). However, for the CIFAR-10 dataset, the \textit{Two EPS} search leads to worse results than the \textit{One EPS} counterpart (see pointer~\rpoint{2}).

\subsection{RoHNAS Results vs. Random Search}

\begin{figure}[h!]
    \centering
    \includegraphics[width=.98\linewidth]{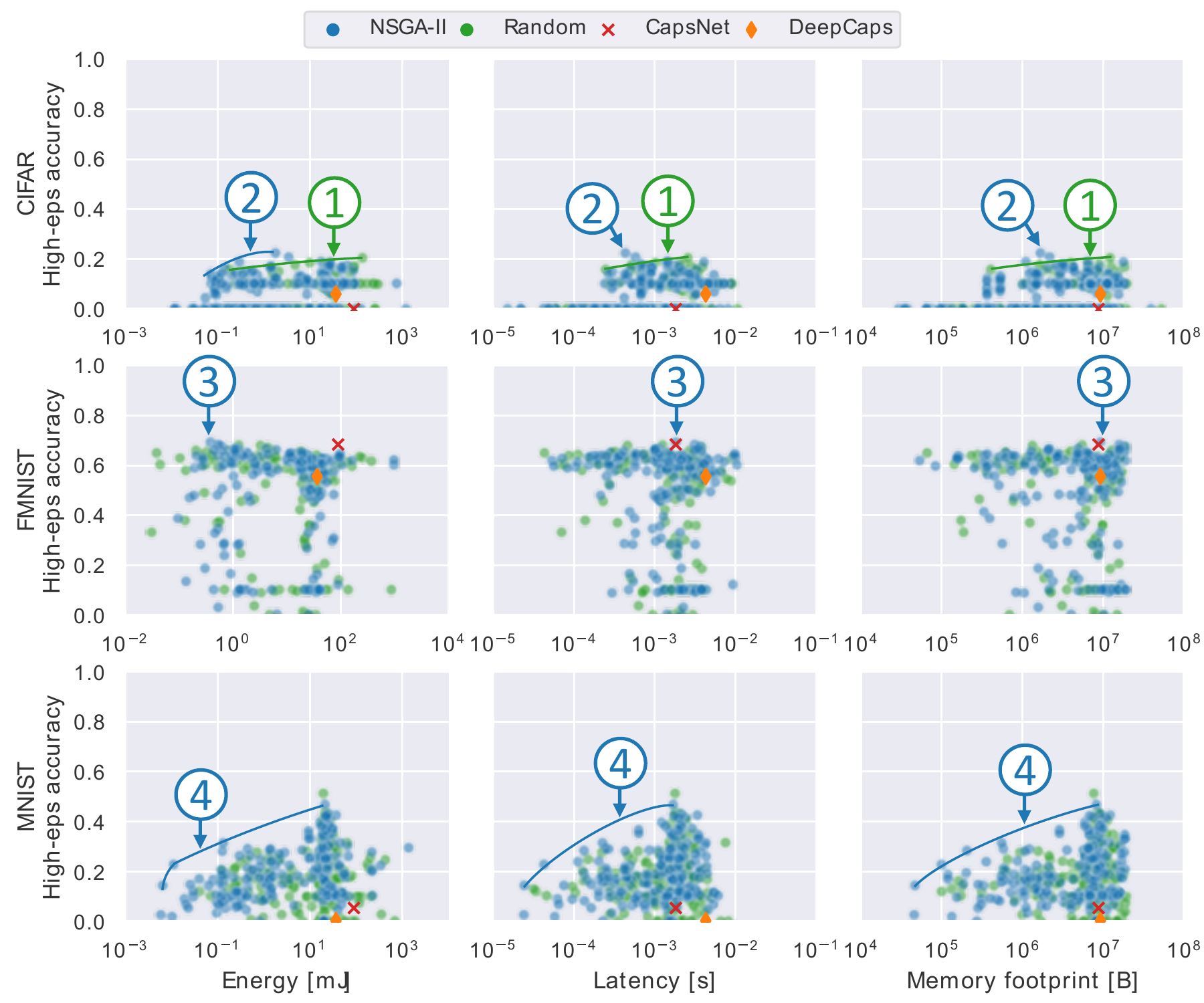}
    \caption{\textit{RoHNAS}' fast evaluation of DNN robustness under PGD attack in the \textit{Two EPS} setting using the $\varepsilon_{high}$ value, compared to the solutions found with random search, showing tradeoffs w.r.t. energy, latency, and memory footprint.}
    \label{fig:joinnas_attack_upper_pointer}
\end{figure}

\begin{figure}[h!]
    \centering
    \includegraphics[width=.98\linewidth]{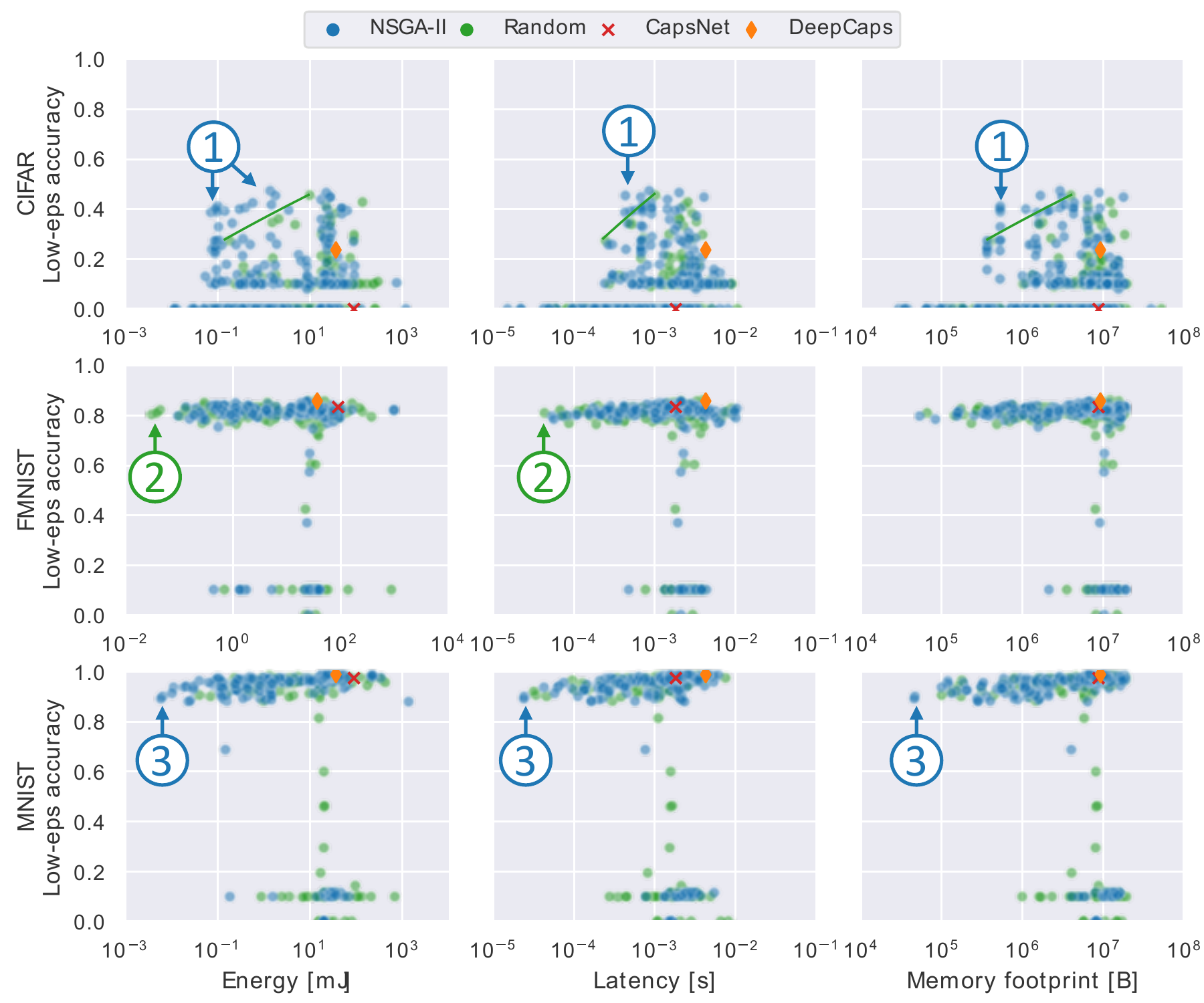}
    \caption{\textit{RoHNAS}' fast evaluation of DNN robustness under PGD attack in the \textit{Two EPS} setting using the $\varepsilon_{low}$ value, compared to the solutions found with random search, showing tradeoffs w.r.t. energy, latency, and memory footprint.}
    \label{fig:joinnas_attack_lower_pointer}
\end{figure}

The \textit{RoHNAS} framework based on the evolutionary search algorithm has been compared to a modified version using random search. The results of the fast DNN robustness evaluation for the \textit{Two EPS} configuration are shown in Figures~\ref{fig:joinnas_attack_upper_pointer} and~\ref{fig:joinnas_attack_lower_pointer}, where Fig.~\ref{fig:joinnas_attack_upper_pointer} shows the accuracy measured when the adversarial perturbation value for the PGD attack is $\varepsilon_{high}$, while Fig.~\ref{fig:joinnas_attack_lower_pointer} uses $\varepsilon_{low}$. Pointer~\rpoint{1} in Fig.~\ref{fig:joinnas_attack_upper_pointer} indicates the Pareto frontier for the CIFAR-10 dataset obtained by random search, which is outperformed by several candidate DNN models found using the NSGA-II algorithm of our \textit{RoHNAS} framework (see pointer~\rpoint{2} in Fig.~\ref{fig:joinnas_attack_upper_pointer}). For the Fashion-MNIST dataset, some candidate DNNs found using the random search show high hardware efficiency, but the solution generated through the NSGA-II algorithm indicated by pointer~\rpoint{3} shows higher robustness. Also for the MNIST dataset, the NSGA-II generates solutions that have better tradeoffs between the objectives, compared to using random search (see pointer~\rpoint{4}).

Similar observations can be made when the adversarial perturbation value for the PGD attack is $\varepsilon_{low}$. As indicated by pointer~\rpoint{1} in Fig.~\ref{fig:joinnas_attack_lower_pointer}, several candidate DNN models for the CIFAR-10 dataset found using the NSGA-II algorithm have better tradeoffs than the Pareto-frontier obtained with random search. However, for the Fashion-MNIST dataset, the solutions with high hardware efficiency (especially low energy and low latency) are found by random search (see pointer~\rpoint{2}). On the other hand, the NSGA-II algorithm generates solutions with higher hardware efficiency for the MNIST dataset (see pointer~\rpoint{3}).

\begin{figure}[h!]
    \centering
    \includegraphics[width=.98\linewidth]{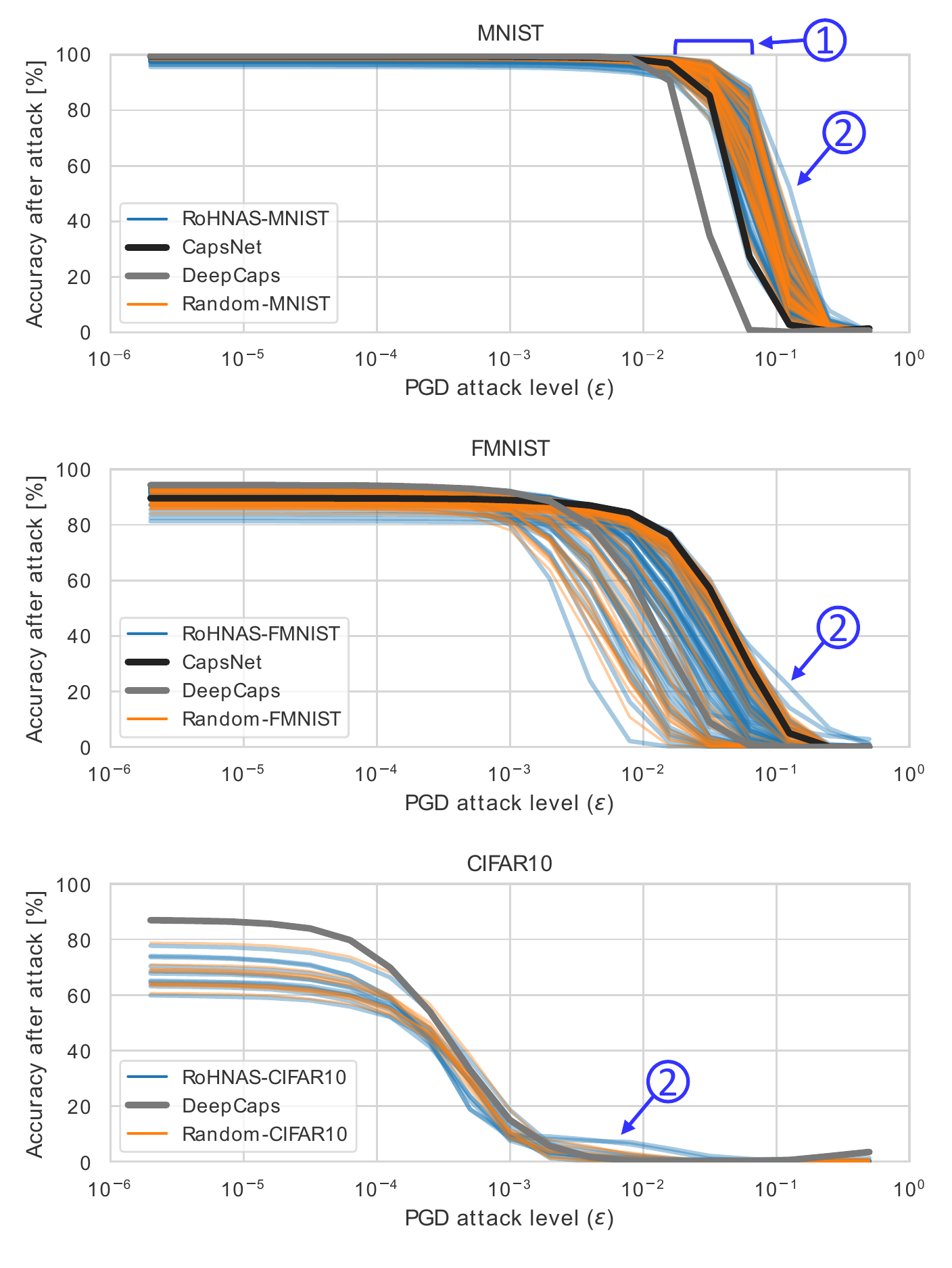}
    \caption{Evaluation of the \textit{RoHNAS} framework with the \textit{Two EPS} setting, compared to the random search.}
    \label{fig:rand_attack_twoeps_final}
\end{figure}

The Pareto-optimal DNNs selected using random search for the \textit{Two EPS} setting are \textit{fully-trained} and compared to the results of the \textit{RoHNAS} framework in Fig.~\ref{fig:rand_attack_twoeps_final}. As highlighted by pointer~\rpoint{1}, the Pareto-optimal solutions generated by both algorithms are robust for a high range of perturbation values $\varepsilon$. However, the key differences can be observed in some curves belonging to the \textit{RoHNAS} search, which exhibit higher robustness than the curves obtained with the random search (see pointer~\rpoint{2} in Fig.~\ref{fig:rand_attack_twoeps_final}).

Our framework supports the integration of different search techniques. Therefore, depending upon the requirements of a system, different search techniques can be run and the best possible solutions can be picked. However, this will lead to a higher experimentation time. Therefore, we recommend using the NSGA-II search algorithm that outperforms the random search in most of the cases.

\section{Conclusion}

In this paper, we proposed \textit{RoHNAS}, a novel framework for the Neural Architecture Search, jointly optimizing for the hardware efficiency (latency, energy, and memory footprint) and robustness against adversarial attacks. Our optimizations for reducing the search space and the exploration time allow finding a set of CNNs and CapsNets, which are Pareto-optimal w.r.t. the above-discussed objectives, in a fast fashion. In our experiments, 900 different DNN models have been evaluated, using 2,000 GPU hours with our fast training settings. Thanks to our \textit{RoHNAS} framework, the deployment of robust DNNs in resource-constrained IoT/neuromorphic edge devices is made possible. We open-source our framework at \url{https://github.com/ehw-fit/rohnas}.



\bibliographystyle{IEEEtran}
\bibliography{main.bib}


\clearpage

\begin{IEEEbiography}[{\includegraphics[width=1in]{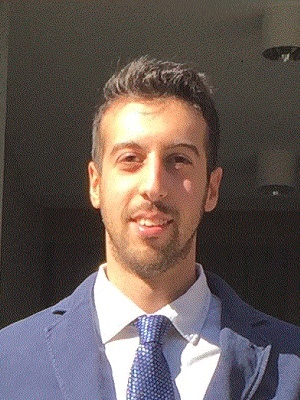}}]{Alberto Marchisio}
(Graduate Student Member, IEEE) received his B.Sc. and M.Sc. degrees in Electronic Engineering from Politecnico di Torino, Turin, Italy, in October 2015 and April 2018, respectively. Currently, he is Ph.D. Student at Computer Architecture and Robust Energy-Efficient Technologies (CARE-Tech.) lab, Institute of Computer Engineering, Technische Universit{\"a}t Wien (TU Wien), Vienna, Austria, under the supervision of Prof. Dr. Muhammad Shafique. His main research interests include hardware and software optimizations for machine learning, brain-inspired computing, VLSI architecture design, emerging computing technologies, robust design, and approximate computing for energy efficiency. He (co-)authored 20+ papers in prestigious international conferences and journals. He received the honorable mention at the Italian National Finals of Maths Olympic Games in 2012, and the Richard Newton Young Fellow Award in 2019.
\end{IEEEbiography}

\begin{IEEEbiography}[{\includegraphics[width=1in]{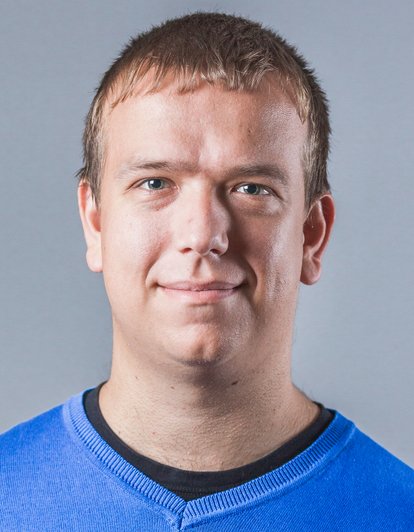}}]{Vojtech Mrazek} (Member, IEEE) received a Ing. and Ph.D. degrees in information technology from the Faculty of Information Technology, Brno University of Technology, Czech Republic, in 2014 and 2018. He is an assistant professor at the Faculty of Information Technology with Evolvable Hardware Group and he was also a visiting post-doc researcher at Department of Informatics, Institute of Computer Engineering, Technische Universit{\"a}t Wien (TU Wien), Vienna, Austria. His research interests are approximate computing, genetic programming and machine learning. He has authored or co-authored over 40 conference/journal papers focused on approximate computing and evolvable hardware. He  received several awards for his research in approximate computing, including the Joseph Fourier Award in 2018 for research in computer science and engineering.
\end{IEEEbiography}

\begin{IEEEbiography}[{\includegraphics[width=1in]{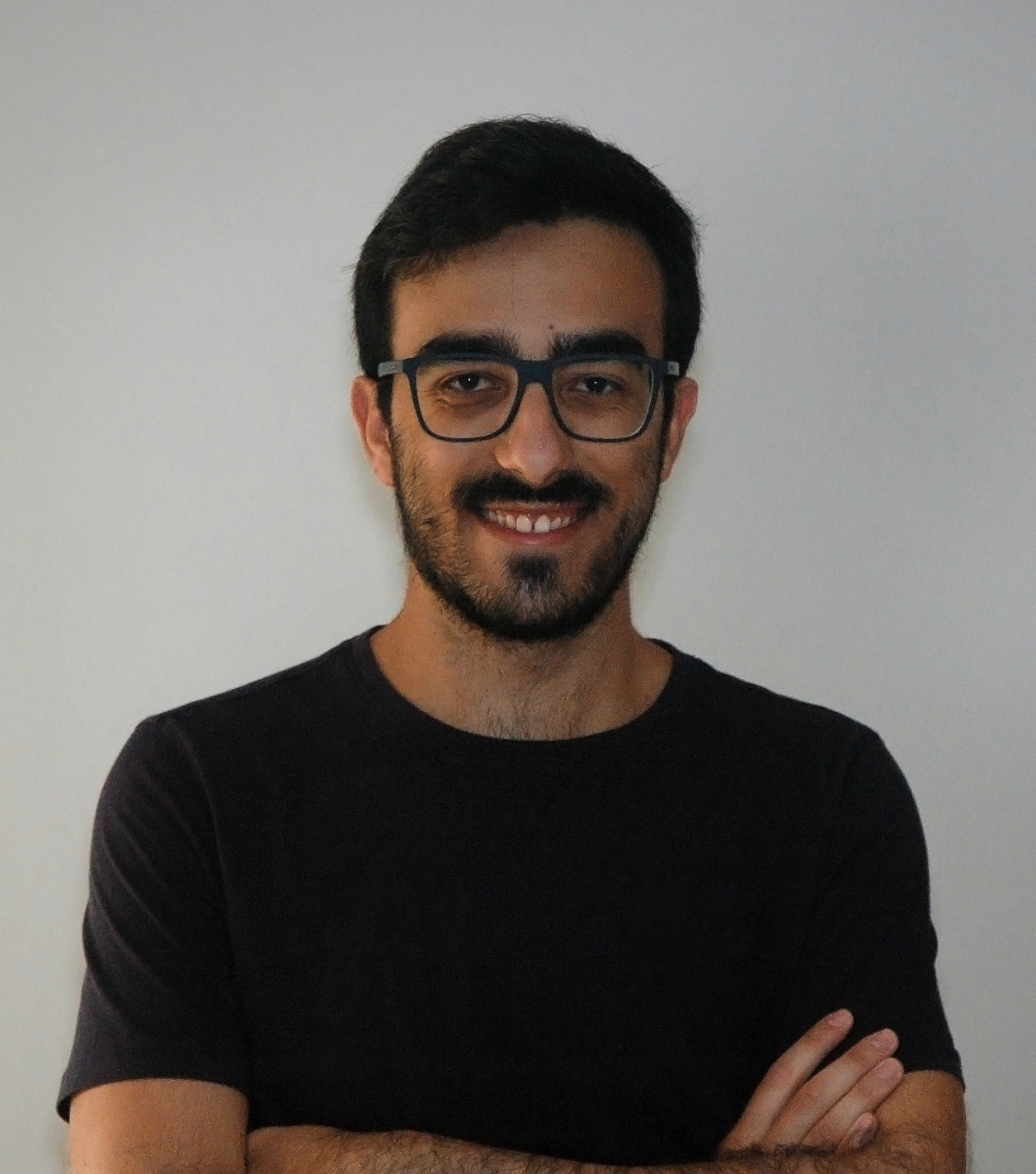}}]{Andrea Massa} received the B.Sc. degree in Electrical and Electronic Engineering in October 2017 from the University of Cagliari, Italy. He received his M.Sc. degree in Electronic Engineering in July 2020 from Politecnico di Torino, Turin, Italy. He is currently a HW and SW Designer in a computer company in Turin. His research interests are in the fields of image processing, computer architecture, machine learning, Deep Neural Networks and genetic algorithms. His previous research work focused on image processing techniques and image registration algorithms applied to real-time automatic camera calibration in the W7-X fusion reactor at the Max Planck Institute for Plasma Physics (Greifswald, Germany) and it was carried out within the activities of the University of Cagliari in different EUROfusion Tasks and Grants. In 2020 his M.Sc. degree thesis work focused on the development of a hardware-aware Neural Architecture Search framework which resulted into a paper that has been accepted for the IEEE/ACM International Conference on Computer-Aided Design (ICCAD 2020).
\end{IEEEbiography}

\begin{IEEEbiography}[{\includegraphics[width=1in]{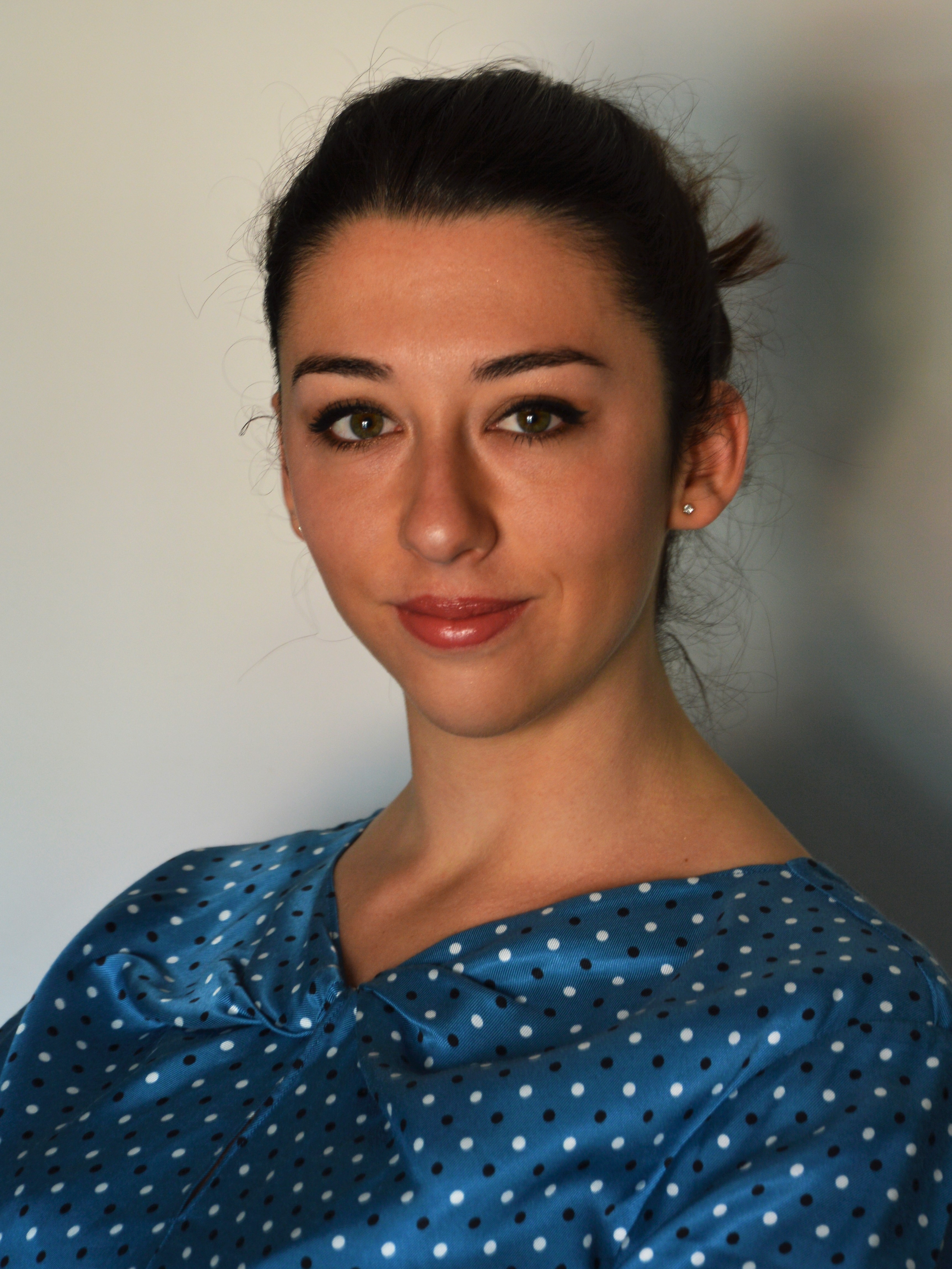}}]{Beatrice Bussolino} (Graduate Student Member, IEEE) received the B.Sc. and M.Sc. degrees in Electronic Engineering, in October 2017 and October 2019 respectively, from Politecnico di Torino, Turin, Italy.  She is now pursuing the Ph.D. degree in Electrical, Electronics and Communications Engineering at Politecnico di Torino under the supervision of Prof. Maurizio Martina. She is an IEEE student member. Her current research interests are in the field of Machine Learning and Deep Neural Networks (DNNs) in particular. The focus of her research activity is the development of on-chip architectures for the edge deployment of DNNs. In 2020, she received the Richard Newton Young Fellow Award and won the DAC Young Fellow Poster Presentation Award.
\end{IEEEbiography}

\begin{IEEEbiography}[{\includegraphics[width=1in]{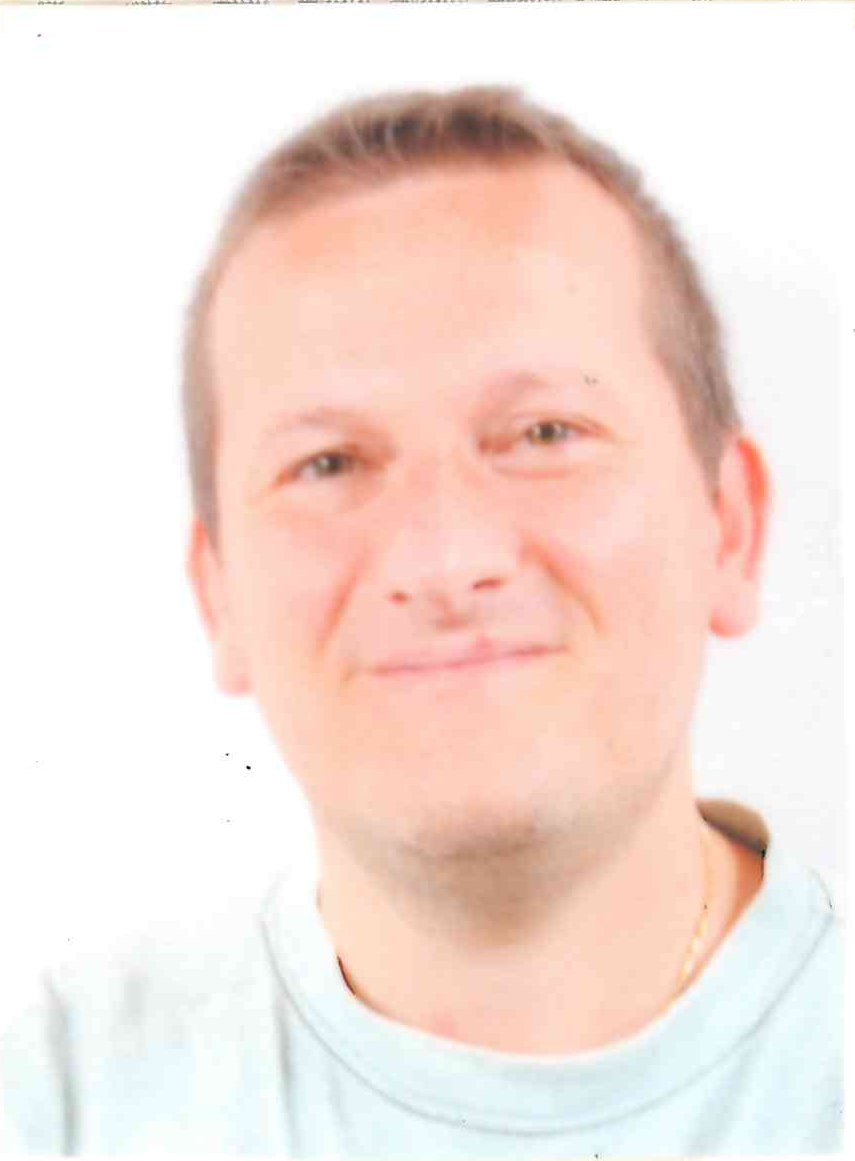}}]{Maurizio Martina} (Senior Member, IEEE) received the M.S. and Ph.D. in electrical
engineering from Politecnico di Torino, Italy, in 2000 and 2004, respectively. He is currently Full Professor with the VLSI-Lab group, Politecnico di Torino. His research interests include computer architecture and VLSI design of architectures for digital signal processing, video coding, communications, networking, artificial intelligence, machine learning and event-based processing. He edited one book and published 3 book chapters on VLSI architectures and digital circuits for video coding, wireless communications and error correcting codes. He has more than 100 scientific publications and is co-author of 2 patents. He is now an Associate Editor of IEEE Transactions on Circuits and Systems - I. He had been part of the organizing and technical committee of several international conferences, including BioCAS 2017, ICECS 2019, AICAS 2020. Currently, he is the counselor of the IEEE Student Branch at Politecnico di Torino and a professional member of IEEE HKN.
\end{IEEEbiography}

\begin{IEEEbiography}[{\includegraphics[width=1in]{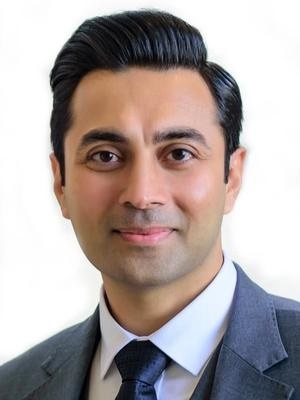}}]{Muhammad Shafique} (Senior Member, IEEE) received the Ph.D. degree in computer science from the Karlsruhe Institute of Technology (KIT), Germany, in 2011. Afterwards, he established and led a highly recognized research group at KIT for several years as well as conducted impactful collaborative R\&D activities across the globe. In Oct.2016, he joined the Institute of Computer Engineering at the Faculty of Informatics, Technische Universität Wien (TU Wien), Vienna, Austria as a Full Professor of Computer Architecture and Robust, Energy-Efficient Technologies. Since Sep.2020, Dr. Shafique is with the New York University (NYU), where he is currently a Full Professor and the director of eBrain Lab at the NYU-Abu Dhabi in UAE, and a Global Network Professor at the Tandon School of Engineering, NYU-New York City in USA. He is also a Co-PI/Investigator in multiple NYUAD Centers, including Center of Artificial Intelligence and Robotics (CAIR), Center of Cyber Security (CCS), Center for InTeractIng urban nEtworkS (CITIES), and Center for Quantum and Topological Systems (CQTS).
His research interests are in AI \& machine learning hardware and system-level design, brain-inspired computing, quantum machine learning, cognitive autonomous systems, wearable healthcare, energy-efficient systems, robust computing, hardware security, emerging technologies, FPGAs, MPSoCs, and embedded systems. His research has a special focus on cross-layer analysis, modeling, design, and optimization of computing and memory systems. The researched technologies and tools are deployed in application use cases from Internet-of-Things (IoT), Smart Cyber-Physical Systems (CPS), and ICT for Development (ICT4D) domains. Dr. Shafique has given several Keynotes, Invited Talks, and Tutorials, as well as organized many special sessions at premier venues. He has served as the PC Chair, General Chair, Track Chair, and PC member for several prestigious IEEE/ACM conferences. Dr. Shafique holds one U.S. patent, and has (co-)authored 6 Books, 10+ Book Chapters, 350+ papers in premier journals and conferences, and 50+ archive articles. He received the 2015 ACM/SIGDA Outstanding New Faculty Award, the AI 2000 Chip Technology Most Influential Scholar Award in 2020 and 2022, the ASPIRE AARE Research Excellence Award in 2021, six gold medals, and several best paper awards and nominations at prestigious conferences. He is a senior member of the IEEE and IEEE Signal Processing Society (SPS), and a member of the ACM, SIGARCH, SIGDA, SIGBED, and HIPEAC.
\end{IEEEbiography}

\EOD

\end{document}